%% file: main.tex
\documentclass[pdflatex,sn-mathphys]{sn-jnl}
\usepackage{array}
\usepackage{comment}
\usepackage{url}
\usepackage{colortbl}
\input{notation_helper.tex}

\begin{document}

\title[A One-Class Classifier for the Detection of GAN Manipulated \dots]{A One-Class Classifier for the Detection of GAN Manipulated Multi-Spectral Satellite Images}

\author*{\fnm{Lydia} \sur{Abady}}\email{lydia.abady@unisi.it}
\author{\fnm{Giovanna Maria} \sur{Dimitri}}\email{giovanna.dimitri@unisi.it}
\author{\fnm{Mauro} \sur{Barni}}\email{barni@dii.unisi.it}

\affil{\orgdiv{Department of Information Engineering and Mathematics}, \orgname{University of Siena}, \orgaddress{\street{via Roma}, \city{Siena}, \postcode{53100}, \state{Siena}, \country{Italy}}}

\abstract{The highly realistic image quality achieved by current image generative models has many academic and industrial applications. To limit the use of such models to benign applications, though, it is necessary that tools to conclusively detect whether an image has been generated synthetically or not are developed. For this reason, several detectors have been developed providing excellent performance in computer vision applications, however, they can not be applied {\em as they are} to multispectral satellite images, and hence new models must be trained. 
In general, two-class classifiers can achieve very good detection accuracies, however they are not able to generalise to image domains and generative models architectures different than those used during training. For this reason, in this paper, we propose a one-class classifier based on \gls{vqvae2} features to overcome the limitations of two-class classifiers. First, we emphasize the generalization problem that binary classifiers suffer from by training and testing an EfficientNet-B4 architecture on multiple multispectral datasets. Then we show that, since the \gls{vqvae2} based classifier is trained only on pristine images, it is able to detect images belonging to different domains and generated by architectures that have not been used during training. Last, we compare the two classifiers head-to-head on the same generated datasets, highlighting the superiori generalization capabilities of the \gls{vqvae2}-based  detector.}

\keywords{Generative Adversarial Networks, Variational AutoEncoder, EfficientNet, Detection, Sentinel-2, Remote sensing.}
\maketitle
\section{Introduction}
\label{sec:intro}

\gls{dl} techniques have established new \gls{sota} benchmarks in several fields: from bioinformatics to computer vision, from natural language processing to object detection \cite{min2017deep,lecun2015deep,otter2020survey,dimitri2022multimodal,zhao2019object}. In this context, the development of tools for the creation of image forgeries, on one hand, and for authenticity verification and other forensic applications, on the other, have seen a steep increase in the use of \gls{dl} techniques \cite{yang2020survey}. Among the possible application domains, great attention is increasingly devoted to satellite image analysis.

As a matter of fact, satellite images play a crucial role in several application areas, such as meteorological forecasts, landscape analysis, agriculture, regional planning, monitoring and detection of natural disasters, and many others. As a result, the number of commercial satellites is constantly growing, and the accessibility of satellite images with larger and large ground resolution \cite{highresasts} is increasing on a daily basis. As for other application domains, \gls{dl} provides several tools to manipulate satellite images. Some examples of \gls{dl}-based tools for satellite images manipulations are described in Ref. \cite{abady2022manipulation} \cite{baier2022synthesis} \cite{zhao2021_cycleganrgb}. Such manipulations are often related to disinformation campaigns, as reported, for instance, in \cite{AussieFire}. Hence, there is a growing need to develop \gls{dl} forensic methods suited for the detection and identification of satellite image forgeries.
The extension of image forensics tools developed for computer vision applications to satellite imagery, however, proves to be challenging from several points of view. First of all, forensic techniques developed for non-satellite images must be adapted to the specific content of overhead imagery. This is due to the inherently different types of features and characteristics of conventional RGB images and multispectral satellite images. In addition, quite often, satellite images have more than 3 bands. For example, Sentinel-2 Level 1C optical images have 13 bands, each group of bands characterized by a different \gls{gsd}. Moreover, unlike RGB images, each pixel is typically represented by more than 8 bits per band (12 bits in the case of Sentinel-2 Level 1C images). On top of that, synthetically generated datasets of multispectral images are missing, thus making it difficult to benchmark \gls{dl} tools for satellite image forgery detection.

\glspl{gan} can be successfully used to create synthetic satellite images \cite{abady2022manipulation}. Such architectures have proven to be extremely useful for image generation or style transfer, and they are applicable not only to RGB images \cite{goodfellow2020generative,isola2017image} but also to multispectral images \cite{abady2020} with only some minor modifications. A few works have also been proposed for the detection of \gls{gan} generated satellite images. In general, most of these tools are trained only on the RGB bands of multispectral images and exhibit good detection capabilities when the training and test datasets are acquired under matched conditions, but they fail to generalize to unseen data. One tool that is actually trained on multispectral images but also lacks generalization capabilities is described in \cite{abady_esann_22}, where the authors present a detector based on an EfficientNet-B4 model trained on multispectral images, achieving very good performance. When the model is tested on a different dataset, though, the performance drop significantly.

To overcome the generalization weakness of \gls{sota} techniques for the detection of \gls{gan} multispectral satellite images, in this work, we propose to use a one-class classifier based on a \gls{vqvae} trained only on pristine images, and to use the reconstruction loss between the input and the output images to distinguish \gls{gan} and pristine images.

We evaluated the performance of the one-class classifier on several multispectral GAN synthetic Sentinel-2 level-1C satellite datasets and compared them against those of a conventional manipulation detector based on EfficientNet-B4, trained on both pristine and \gls{gan}-generated satellite images. We run the experiments on the full 13 bands of Sentinel-2 level-1C samples. The results we got, demonstrate the superior performance of the one-class classifier in terms of generalization capability, with only a small performance loss with respect to the two-class classifier under matched conditions.

The paper is structured as follows: in Section \ref{sec:sota}, we overview the state of the art in the field of satellite imagery forgery detection. In Section \ref{sec:datasets}, we describe the datasets we created or collected, differentiating between the various models and \glspl{gan} methodologies used. In Section \ref{sec:vqvae2}, we describe the \gls{vqvae} one-class classifier, and in Section \ref{sec:eval}, we present the experimental results proving the validity of the proposed method. Eventually, Section \ref{sec:conc} draws conclusions and offers future perspectives on our work.

\section{State of the Art}
\label{sec:sota}

In this section, we overview the prior art on satellite images forgery detection. In \cite{yarlagadda2018satellite}, the authors introduce a framework composed of two steps for the detection and localization of forgeries in satellite images. In the first step, a \gls{gan} is trained to obtain a set of features capable of representing pristine satellite images. In the second step, a one-class classifier based on \gls{svm} is applied to distinguish pristine and non-pristine images. 
The method that is proposed in \cite{bartusiak2019splicing} is based on a conditional \gls{gan} architecture that is trained on two domains, the first domain is the domain of spliced images while the second contains the forgery masks. Given a \gls{gan} image as input, the \gls{gan} generator is used to estimate a forged mask that is as close as possible to the real one. In \cite{horvath2019anomaly}, the authors apply an analysis similar to \cite{yarlagadda2018satellite}, by jointly training the auto-encoder and a \gls{svdd} \cite{tax_svdd_2004}. Furthermore, in \cite{horvath2020manipulation}, the authors take advantage of a one-class \gls{dbn}, which is employed for detecting and localizing forged images. The same authors, in \cite{horvath2021_visiontransformer4detection}, propose to use a Vision Transformer for image reconstruction. In this case, the forgery mask is obtained by observing the differences between the input and output images.

Another interesting work is \cite{horvath2021_attentionunet}, where the heat maps of forged regions are estimated by using a novel architecture based on a U-Net nested within a GAN architecture. The system built in this way is able to localize RGB image forgeries generated by 3 different types of GANs: StyleGAN2 \cite{karras_2019}, ProGAN \cite{karras_2017} and CycleGAN \cite{zhu_2017}. The forgeries are created by using Sentinel-2 RGB images that are spliced within the generated images. The output of the architecture consists of a probability mask, where each pixel is associated with the probability of having been generated by one of the specific GANs that are tested in the experimental framework.

In \cite{zhao2021deep}, the authors implement a \gls{gan}-based approach for semantic satellite image translation. In the same work, they also present a data-driven approach for the detection of \gls{gan} generated images and use \gls{svm} to detect cycleGAN-generated images from a set of features (both spatial and spectral). In \cite{chen2021geo}, the authors rely on the dataset presented in \cite{zhao2019object}, to develop an additional method to detect RGB satellite images that are semantically transformed, starting from the detection of high-frequency details in the generated samples.

Most of the methods proposed so far focus on RGB 8-bit images. To the best of our knowledge, the only method that has been proposed to detect multispectral images forgeries on images with 13 bands, like those acquired by Sentinel-2 sensors, is \cite{abady_esann_22}, which, can not generalize to mismatched data, as we already mentioned in the Introduction.

\section{Datasets}
\label{sec:datasets}

In this section, we describe the datasets that we have used to train the detectors and test them. The datasets consist of two main types of images: \emph{pristine} Sentinel-2 level1-C images and \emph{\gls{gan}} images generated from models trained on Sentinel-2 level1-C datasets.
The pristine images were obtained from the ESA Copernicus hub \cite{esa_2021}. In our datasets, the images consist of 13 bands. The spatial resolution of the green, blue, red, and \gls{nir} bands (bands 2, 3, 4, and 8 respectively) is 10m, with an overall size of 10980 $\times$10980 pixels. Bands 5, 6, 7, 8a, 11, and 12, instead, have 20m spatial resolution and a size of 5490$\times$5490 pixels. Finally, bands 1, 9, and 10 have a spatial resolution of 60m, and size equal to 1830$\times$1830 pixels. The radiometric resolution of all of the bands is 16 bits.

Prior to their use, all the bands that did not have a resolution of 10m were up-sampled. This allowed us to deal with images having the same size of the 10m image bands. We then divided the images into 512$\times$512 patches. Tiling was implemented by using the gdal-retile of the gdal software library \cite{gdal_2020}. Moreover, as a further pre-processing step, we removed from all the datasets the tiles with no data pixels (0 brightness).

For the generalization experiments, we collected an additional dataset, hereafter referred to as "This-city-does-not-exist" dataset, containing RGB images of size 512$\times$512. A detailed description of all the datasets is given in the next subsections.

A summary of all the datasets described in this section is given in Table \ref{tab:datasets}.

\begin{table}[htb]
\centering
\resizebox{\textwidth}{!}{%
\begin{tabular}{|l|l|l|l|l|l|l|}
\hline
\rowcolor{lightgray}
\rule{0pt}{15pt}
\textbf{ Dataset} & \textbf{ Bands}  & \textbf{ Size} & \textbf{ Architecture} & \textbf{ Transfer Type} & \textbf{ total \# pristine}& \textbf{ total\# \gls{gan}}\\
\hline
\rule{0pt}{10pt}
\textbf{Land Cover (LC)}     & 13 & 512$\times$512             & CycleGAN              & Land Cover        & 30000  & 4000        \\ \hline
\rule{0pt}{10pt}
\textbf{Scandinavian (Scand)} & 13  & 512$\times$512             & Pix2pix               & Season         & 17044 & 4000           \\ \hline
\rule{0pt}{10pt}
\textbf{China}               & 13  & 512$\times$512             & Pix2pix               & Season         & 16000 & 4000          \\ \hline
\rule{0pt}{10pt}
\textbf{Alps}           &      13    & 512$\times$512           &               --        &               --     & 7872 & 0      \\\hline
\rule{0pt}{10pt}
\textbf{This city does not exit}           &      3    & 1024$\times$1024           &               styleGAN2        &               --     & 0 & 140      \\ \hline
\end{tabular}}

\caption{Summary of the datasets used in our work (see Section \ref{sec:datasets}). The first column reports the names of the datasets. The other columns report the number of bands, the size, the type of architecture used to generate the images, the type of transfer applied to the images, the total number of pristine images and the total number of \gls{gan} images.}
\label{tab:datasets}
\end{table}

The datasets were used for several tasks. To start with, we needed a set of pristine images to train the \glspl{gan} architectures and the \gls{vqvae2}-based detector, the pristine images and the \gls{gan} images were used to train the 2-class EfficientNet detector, in addition the \gls{gan} images were used to test both detectors. Eventually, a small number of pristine images were used to calibrate the thresholds for both detectors since the assessment metric we used was the probability of detection at a false alarm rate of 0.1. Table \ref{tab:datasets_tasks} shows how we split the images of the various datasets across different tasks.

\begin{table}[htb]
\resizebox{\textwidth}{!}{%
\begin{tabular}{|c|c|c|c|c|c|}
\hline
\rowcolor{lightgray}
\textbf{Dataset} & \textbf{Train GAN} & \textbf{Train EfficientNet-B4} & \textbf{Train VQ VAE 2} & \textbf{Test Detectors} & \textbf{Calibrate threshold of detectors} \\ \hline
\textbf{LC}      & 16000 P            & 3000 P, 3000 G              & 29000 P                 & 1000 G            & 100 P                              \\ \hline
\textbf{Scand}   & 13044 P            & 3000 P, 3000 G              & --                      & 1000 G            & 100 P                              \\ \hline
\textbf{China}   & 12000 P            & 3000 P, 3000 G              & 15000 P                 & 1000 G            & 100 P                              \\ \hline
\textbf{Alps}    & --                 & --                          & 7872 P                  & --                      & --                                        \\ \hline
\textbf{This-city-does-not-exist}    & --                 & --                          & --                  & 140 G                      & --                                        \\ \hline
\end{tabular}}
\caption{Summary of the number of images used for each task. P indicates Pristine images and G denotes \gls{gan} generated images. For \gls{vqvae2} training, all the images were used to train a single one-class detector. For EfficientNet-B4, the training images were used to train several versions of the detectors each time by using a different combination of the training sets (see Section \ref{ssec:eff_eval_detect}). The images used for testing were never used during training}
\label{tab:datasets_tasks}
\end{table}

\begin{table}[htbp]
\centering
\resizebox{\textwidth}{!}{%
\begin{tabular}{cccccc}
\includegraphics[width=0.16\columnwidth]{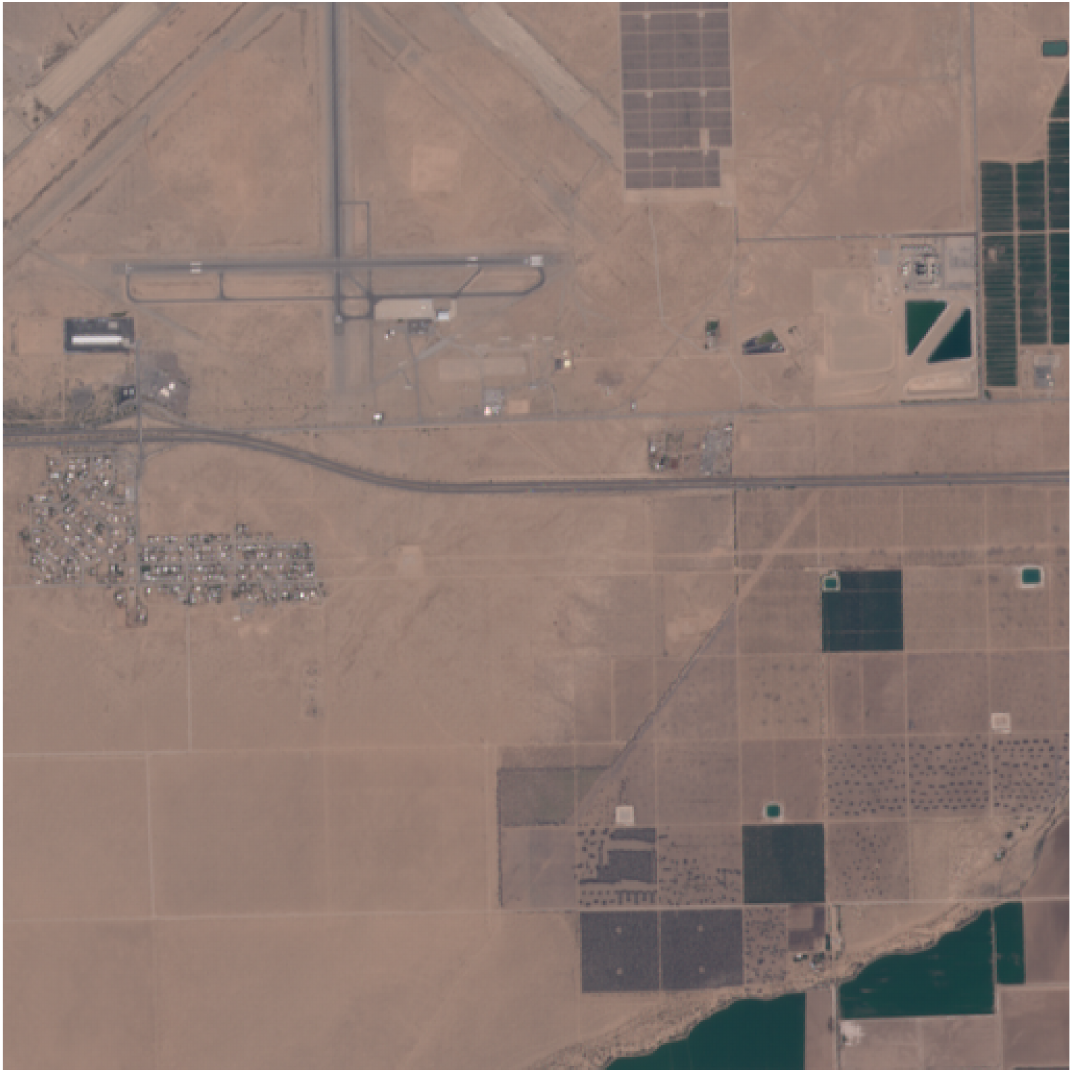}
&\includegraphics[width=0.16\columnwidth]{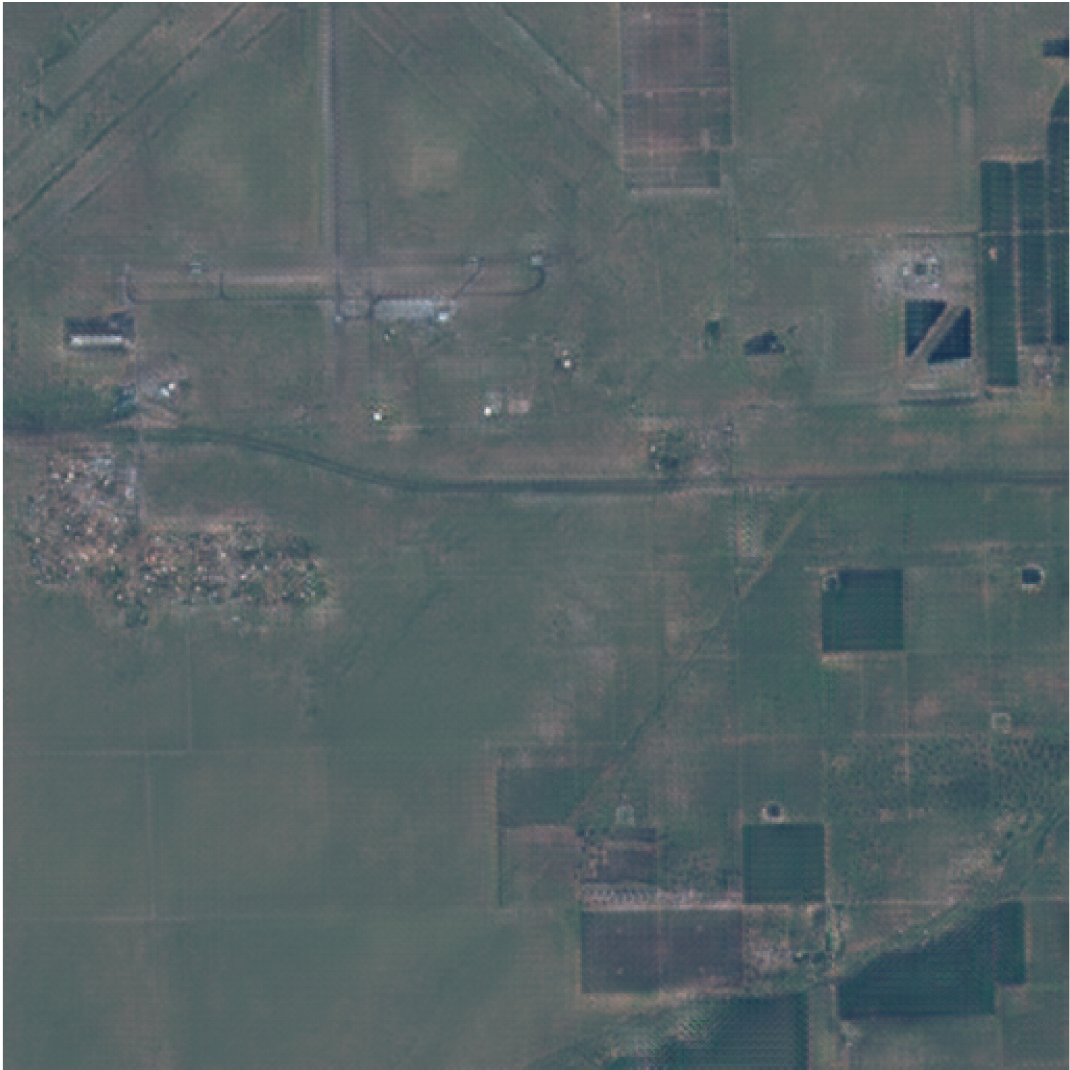}
&\includegraphics[width=0.16\columnwidth]{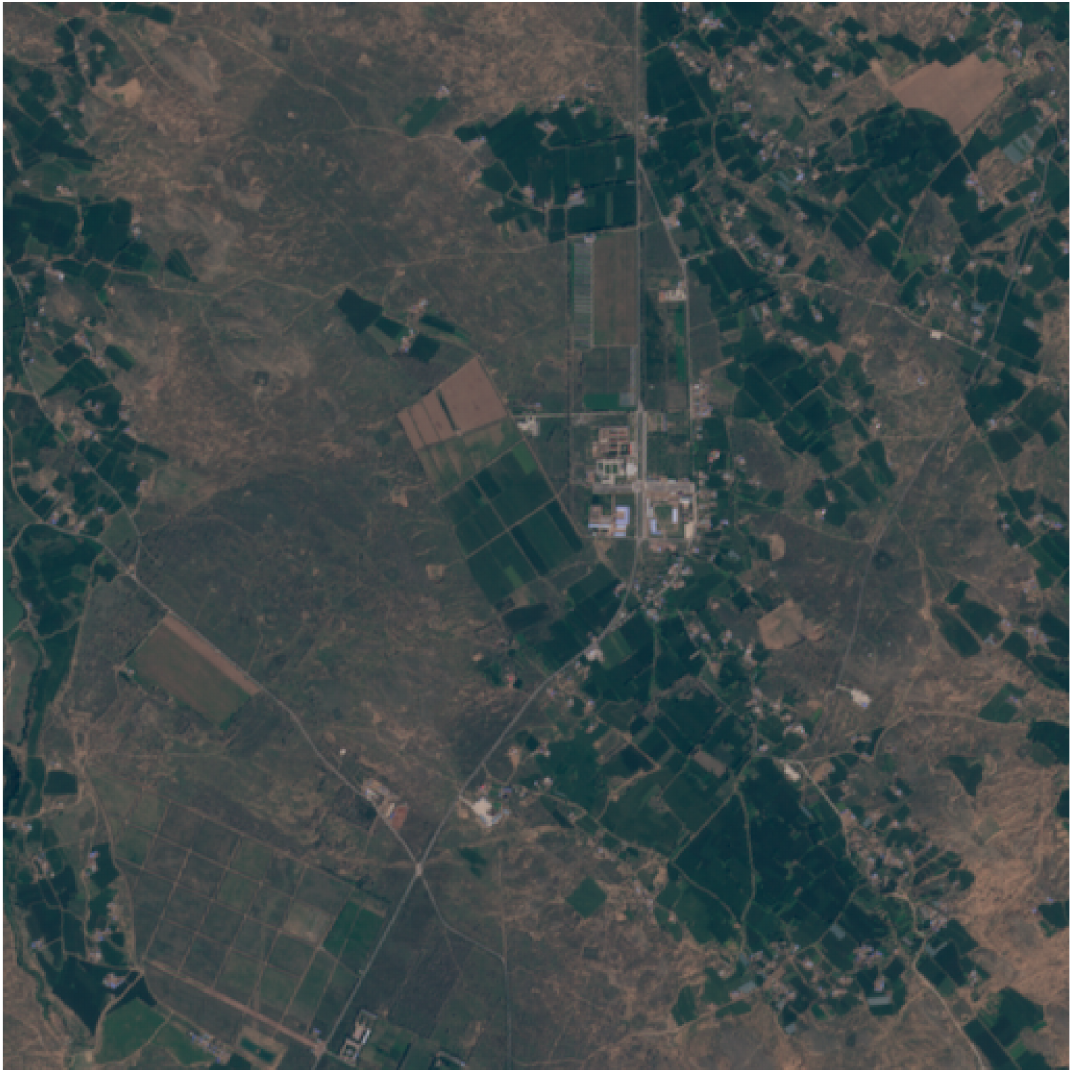}
&\includegraphics[width=0.16\columnwidth]{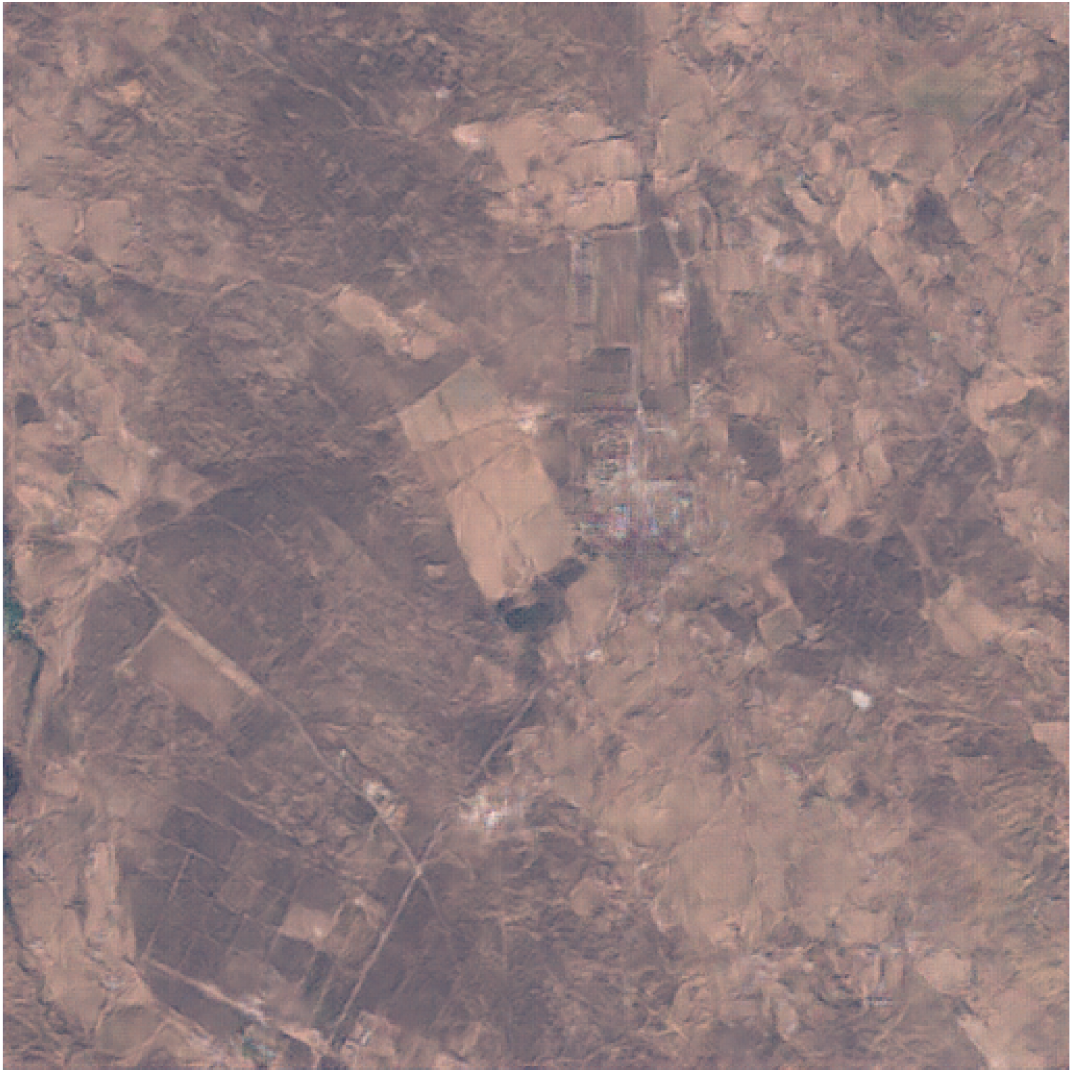}
&\includegraphics[width=0.16\columnwidth]{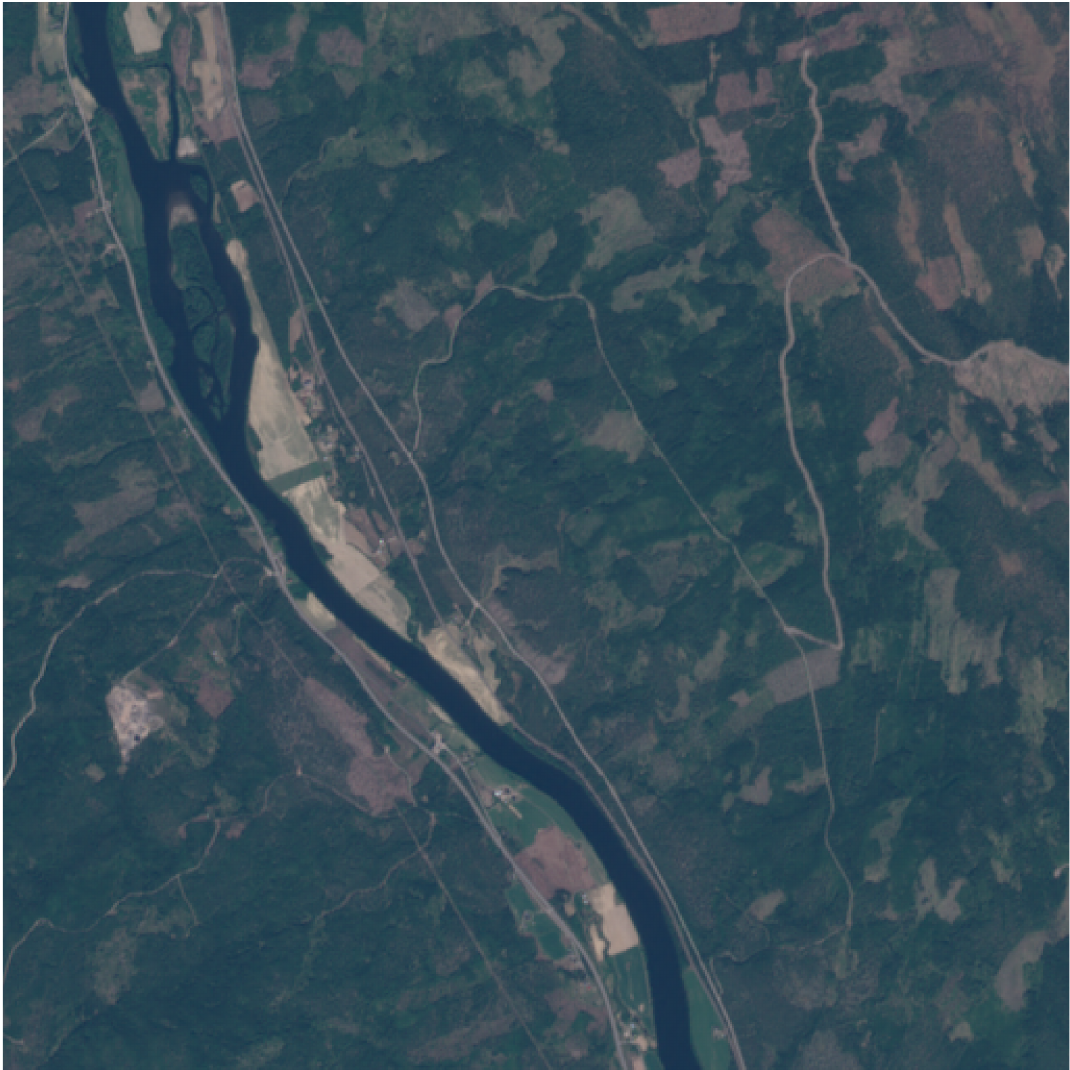}
&\includegraphics[width=0.16\columnwidth]{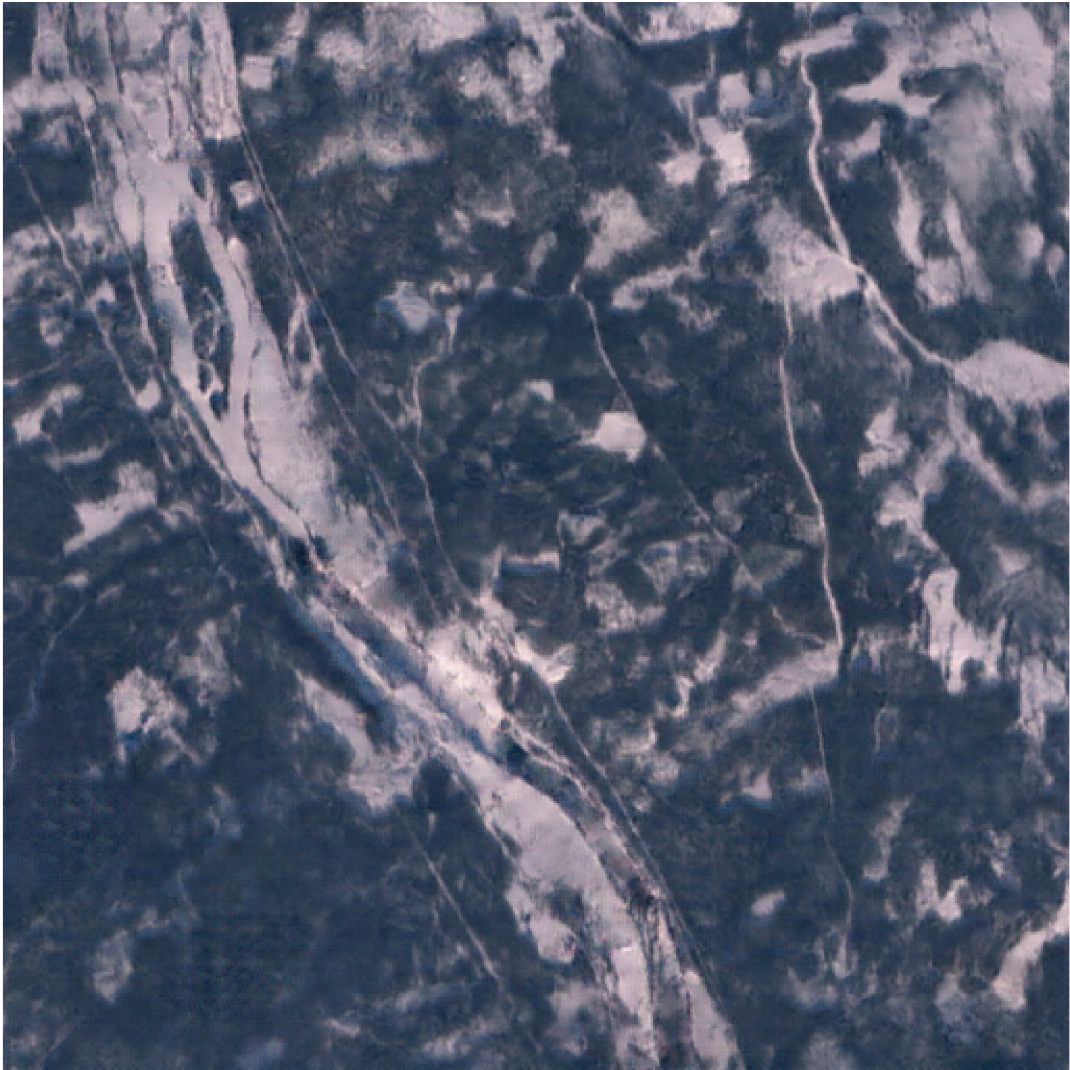}
\\
\multicolumn{2}{c}{Barren to Vegetation}&
\multicolumn{2}{c}{Summer to Winter}&
\multicolumn{2}{c}{Summer to Winter}
\\
\includegraphics[width=0.16\columnwidth]{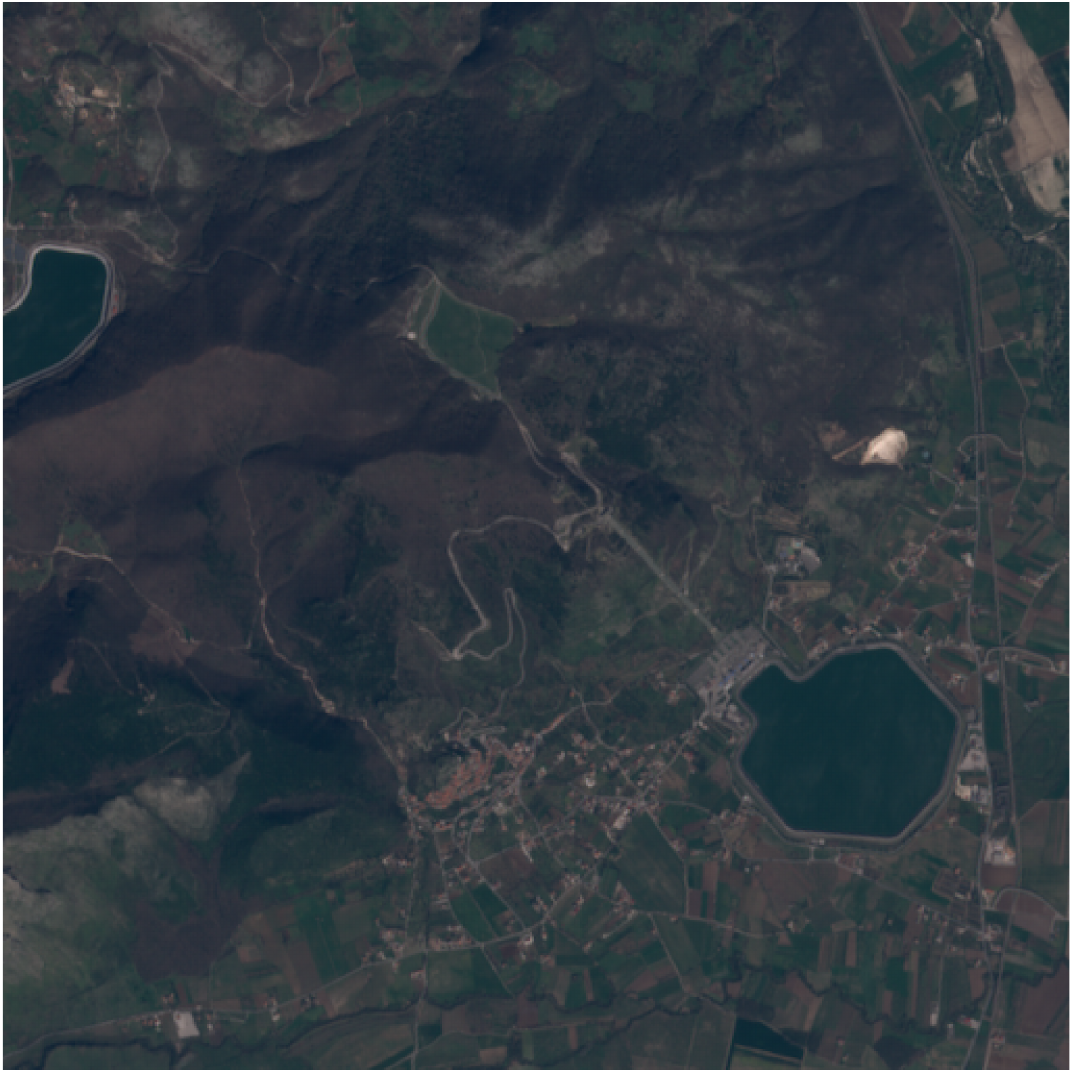}
&\includegraphics[width=0.16\columnwidth]{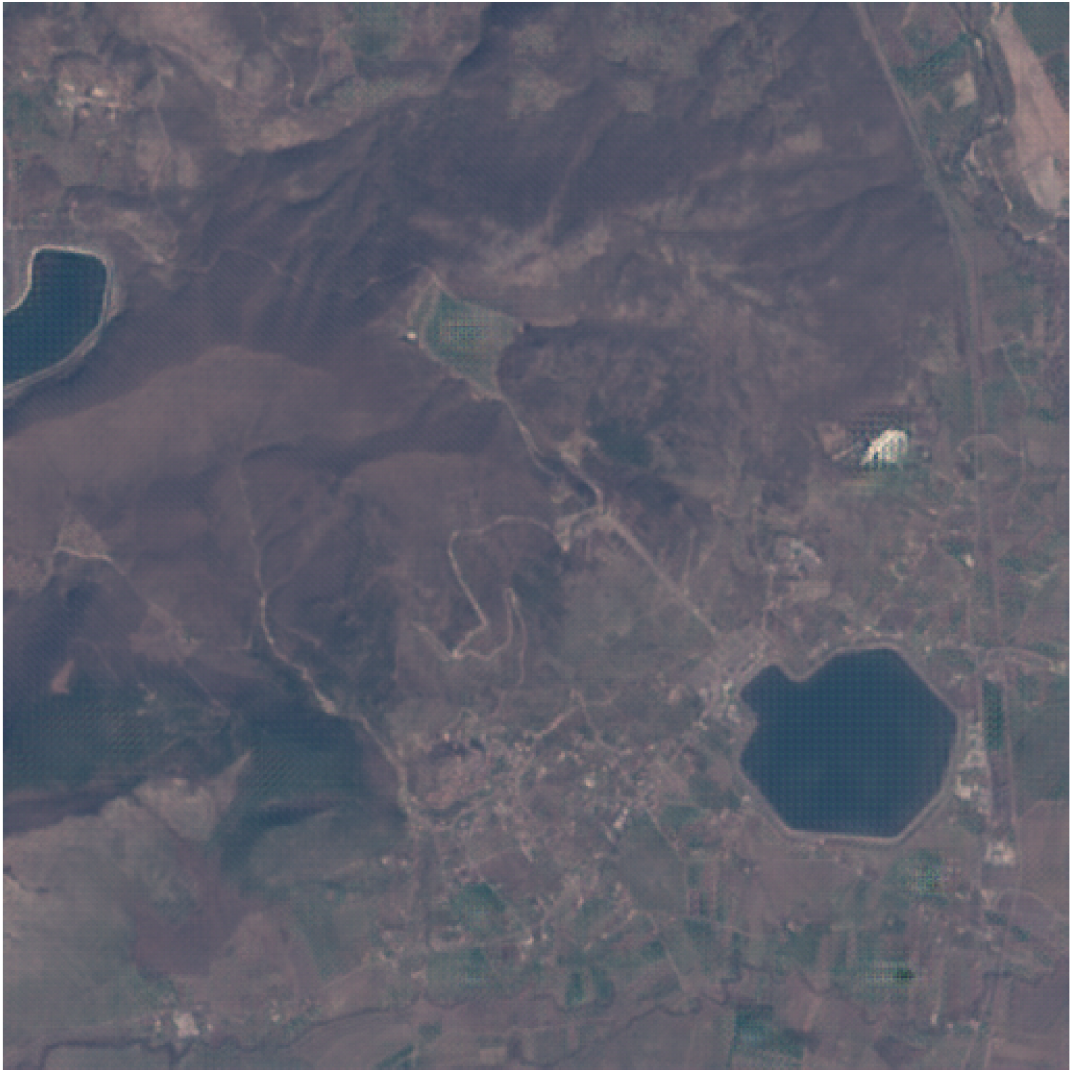}
&\includegraphics[width=0.16\columnwidth]{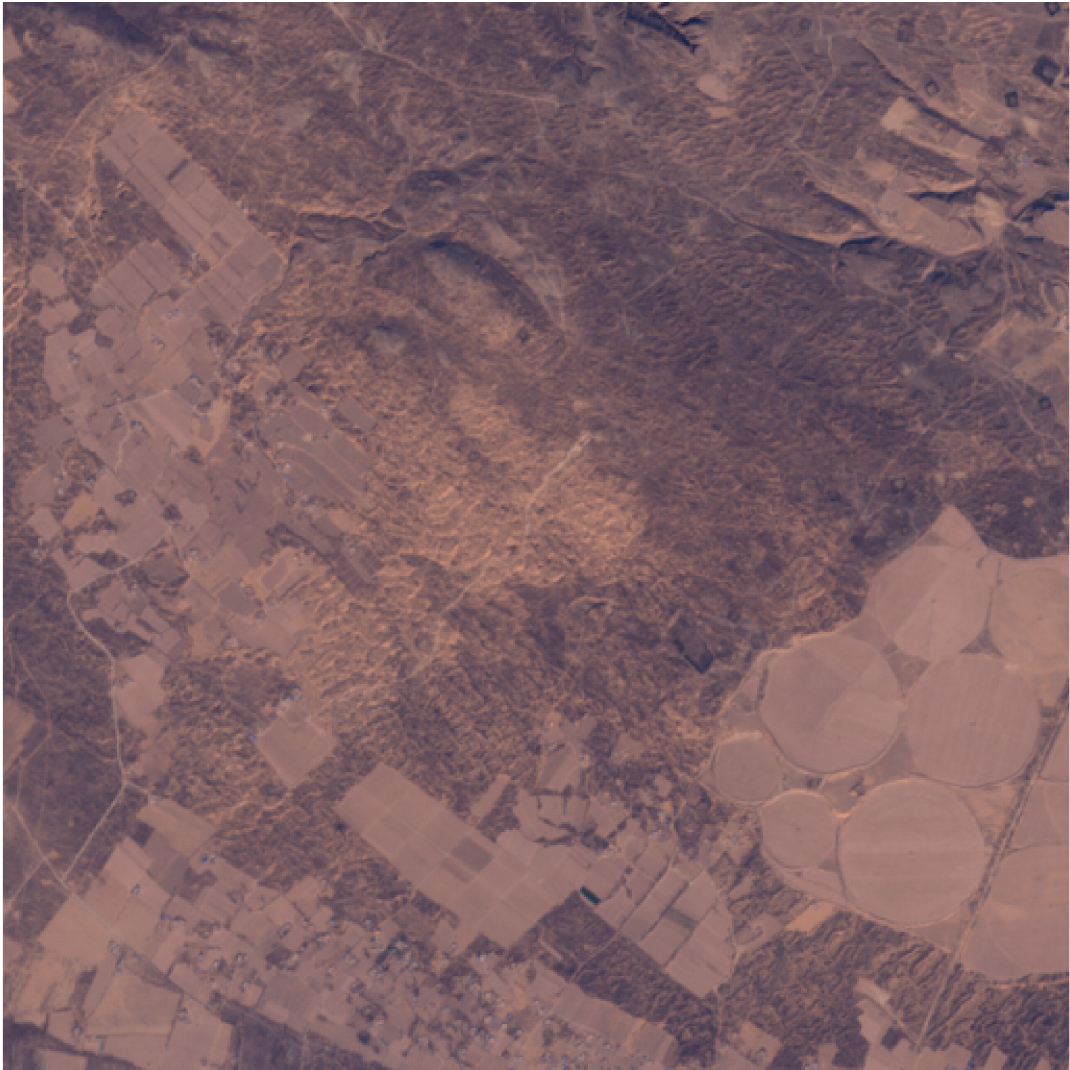}
&\includegraphics[width=0.16\columnwidth]{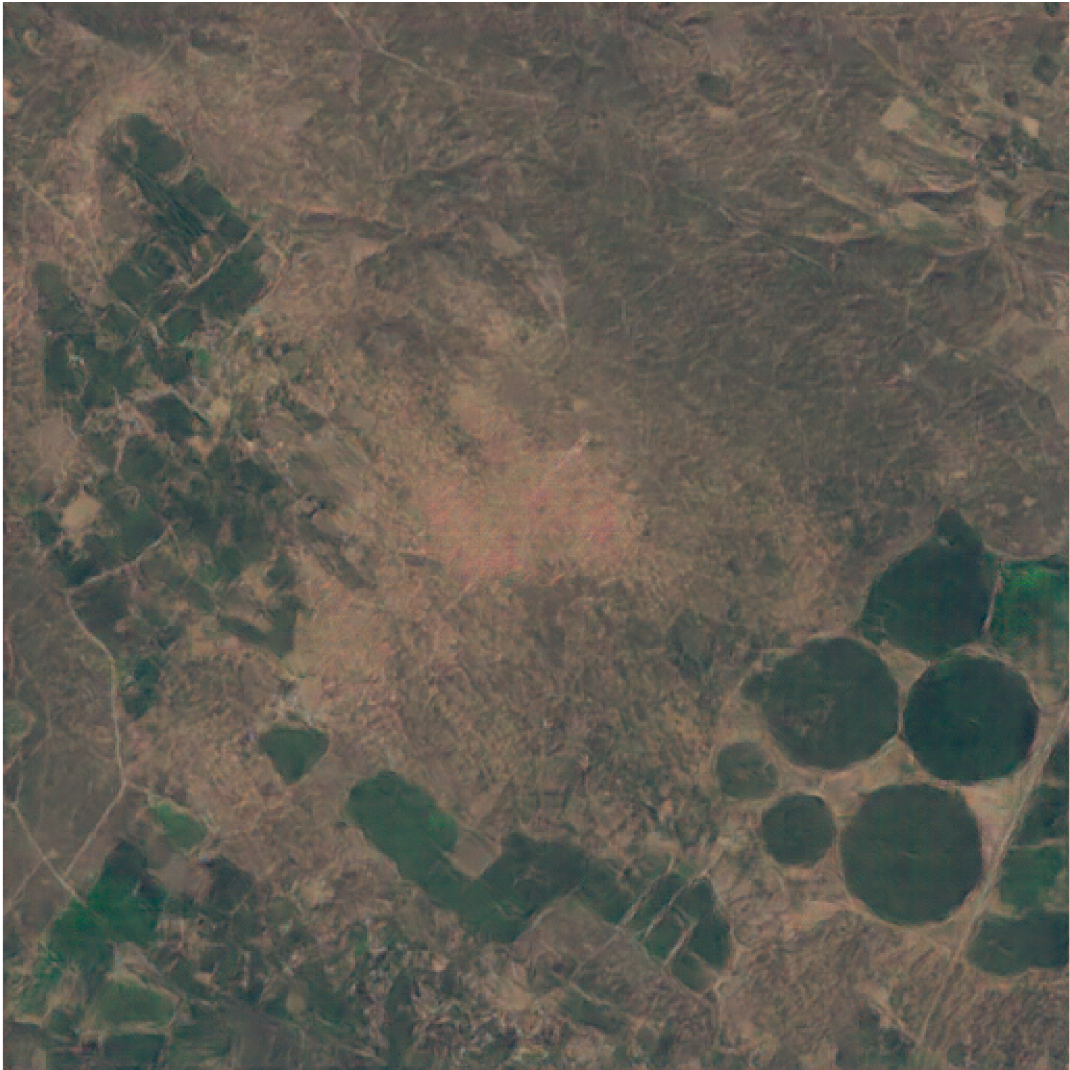}
&\includegraphics[width=0.16\columnwidth]{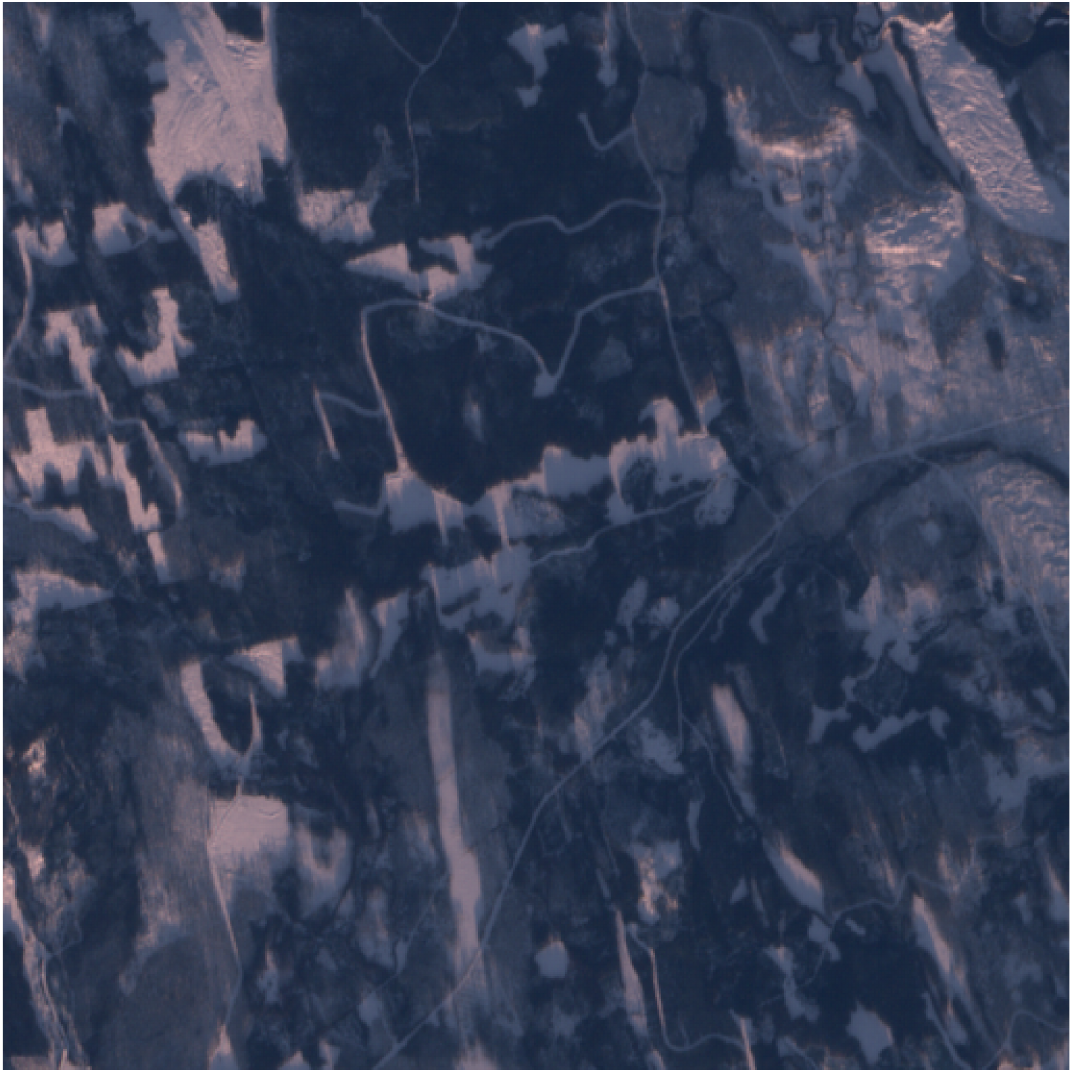}
&\includegraphics[width=0.16\columnwidth]{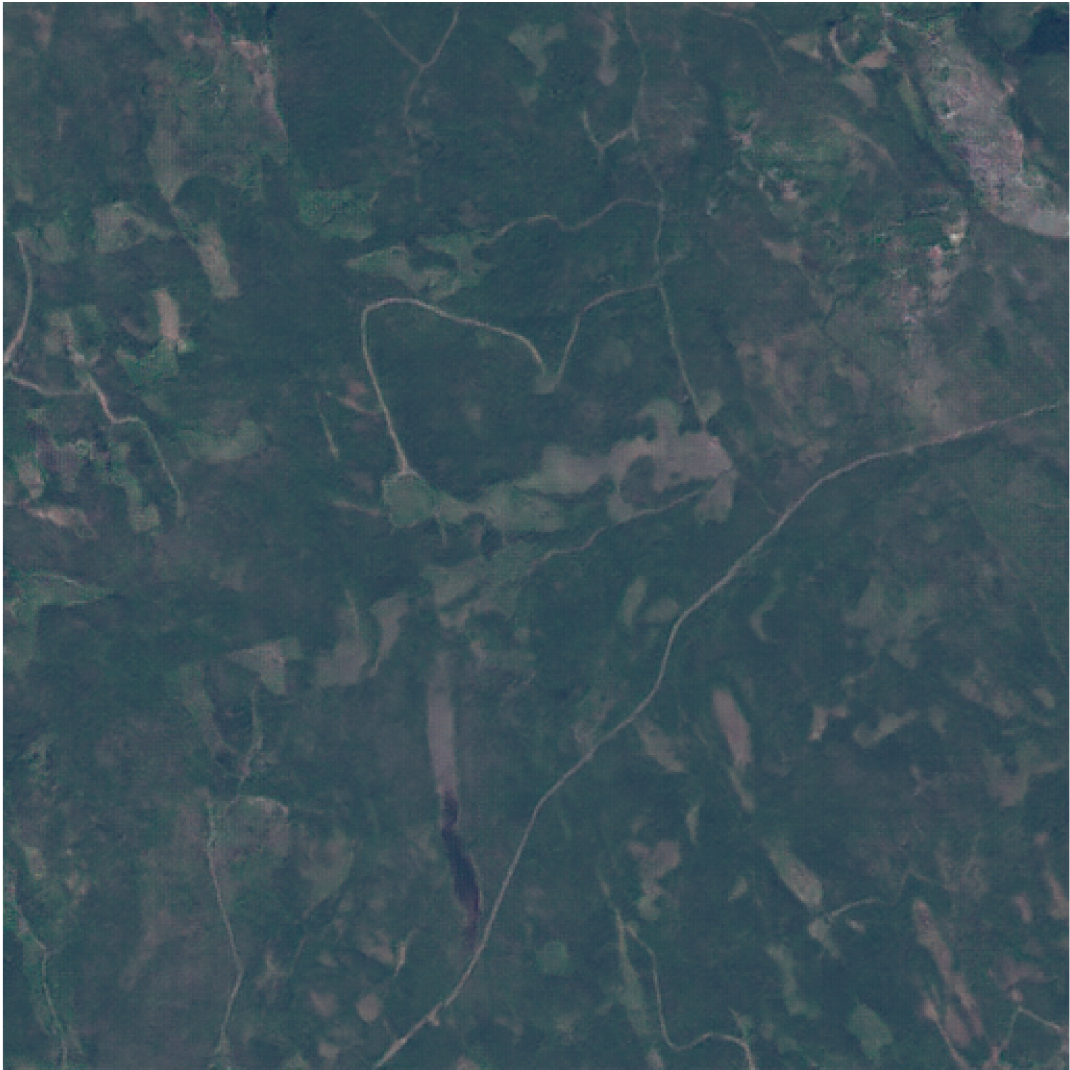}
\\
\multicolumn{2}{c}{Vegetation to Barren}&
\multicolumn{2}{c}{Winter to Summer}&
\multicolumn{2}{c}{Winter to Summer}
\\

{(a) LC - Pristine}&{(b) LC - GAN}&
{(c) China -  Pristine}&{(d) China -  GAN}&
{(e) Scand -  Pristine}&{(f) Scand -  GAN}
\\
\end{tabular}}
\captionof{figure}{some examples of the images contained in the datasets used throughout the paper. Only the RGB bands are shown.}
\label{fig:datasets_examples}
\end{table}

\subsection{Land Cover (LC) Transfer Datasets}
\label{ssec:lc_datasets}
The first dataset we built is the land cover (LC) transfer dataset. The dataset is formed by two classes of images: pristine and \gls{gan} images. To generate the \gls{gan} images, we first gathered the pristine images that were used to train the cycleGAN architecture \cite{zhu_2017} in charge of transferring the land cover from one domain to another. As indicated in Table \ref{tab:datasets_tasks}, the pristine images were also used to train the \gls{gan} images detectors. The cycleGAN was instructed to transfer barren to vegetation landscapes and vice versa. For this reason, the dataset had to contain two kinds of images: images dominated by vegetation, and images mostly containing barren terrain. To build the dataset, we exploited the organization for economic cooperation and development land cover classifications data~\cite{oecd_2020}. For the vegetation domain, we obtained images from the following countries: Salvador, Congo, Montenegro, Gabon, and Guyana. For the barren domain, we downloaded images from South and Central America. In total, we gathered 10000, 512$\times$512 images per domain (vegetation and barren). The cycleGAN was trained on 16000 pristine images equally split into vegetation and barren terrains. The remaining 4000 images, were used to generate 4000 \gls{gan} images using the trained cycleGAN model. We relied on this set to train the efficientNet-B4 detector on 3000 pristine images and 3000 generated images. The remaining 1000 \gls{gan} images were used for testing, while 100 pristine images were used for threshold calibration. To train the \gls{vqvae2}, gathered 10000 additional pristine images (5000 barren and 5000 vegetation)\footnote{Then, in total, the LC dataset contains 30000 pristine images.}. Then, we used 29000 pristine images (all the images except 1000 images we left aside) as part of the \gls{vqvae2} training dataset. Figure \ref{fig:datasets_examples}a shows two examples from the pristine LC dataset. Figure \ref{fig:datasets_examples}b, instead, shows two examples of images generated by the cycleGAN.

\subsection{China and Scandinavian Season Transfer Datasets}

The China and Scandinavian (Scand) Season Transfer datasets are another example of image-to-image \gls{gan} image generation. Instead of changing the type of land cover, in this case the network was asked to generate the summer (res. winter) counterpart of a winter (res. summer) image. To create these datasets, we started from images taken at two different geographic locations (China and Scandinavia) characterized by very different seasonal changes.
To generate the \gls{gan} images of the China season transfer dataset, we trained two pix2pix \gls{gan} models \cite{isola_2016}. Since pix2pix is an architecture for one-way style transfer, we needed to train two different models to achieve a season-style transfer in both directions. For training, we needed a paired dataset, so we decided to download images taken in China at the same location but in two different months of the year. Specifically, we retrieved images taken in August 2020 and in January 2021. The total number of pristine images collected for the China dataset is 16000, forming 8000 pairs out of which we took the 6000 pairs we used to train the \glspl{gan}. We then created a dataset from the 13-band \gls{gan} models that contained 8000 images, equally divided into pristine and \glspl{gan} and equally divided into summer and winter. 6000 images out of these, were used to train the efficientNet-B4 detector, 100 pristine images were used for threshold calibration, and 1000 \gls{gan} images were used to test the two detectors. In addition, 15000 pristine images from this dataset (excluding the images used for calibration and testing) were used as part of the \gls{vqvae2} training dataset. Figure \ref{fig:datasets_examples}c shows two examples of pristine images, while Figure \ref{fig:datasets_examples}d shows two examples of \gls{gan} generated images.

The Scandinavian season transfer dataset was created in a similar manner, with summer images taken in June 2020 and winter images in February 2020. See Figures \ref{fig:datasets_examples}e and \ref{fig:datasets_examples}f for two examples of pristine and \gls{gan} images. The total number of pristine images collected for this dataset is 17044, which corresponds to 8522 pairs. The style transfer \gls{gan}s were trained on 6522 pristine image pairs, while the efficientNet-B4 detector was trained on 6000 images, the threshold calibration of the detectors was performed on 100 pristine images, and the tests were carried out on 1000 \gls{gan} images.

\subsection{Alps Dataset}
The {\em Alps} dataset is a pristine image dataset collected to help training the \gls{vqvae2} detector. We collected images from the exact same area in two different months, with each month representing a different season (June 2019 for summer and December 2019 for winter). To avoid generating images with clouds, we selected images with limited cloud cover. Since it was not possible to obtain images with 0\% cloud cover, we limited the search to images with cloud cover less than 9\%. As a result, we obtained a dataset with 7872 pristine images. Figure \ref{fig:citydoesntexist_alps}-b shows an RGB representation of two examples of the Alps dataset.

\subsection{This-city-does-not-exist Dataset}
\label{ssec:city_dataset}
\begin{figure}[htpb]
\centering

\begin{subfigure}{.48\columnwidth}
\label{fig:citydoesntexist}
\centering
\includegraphics[width=0.49\columnwidth]{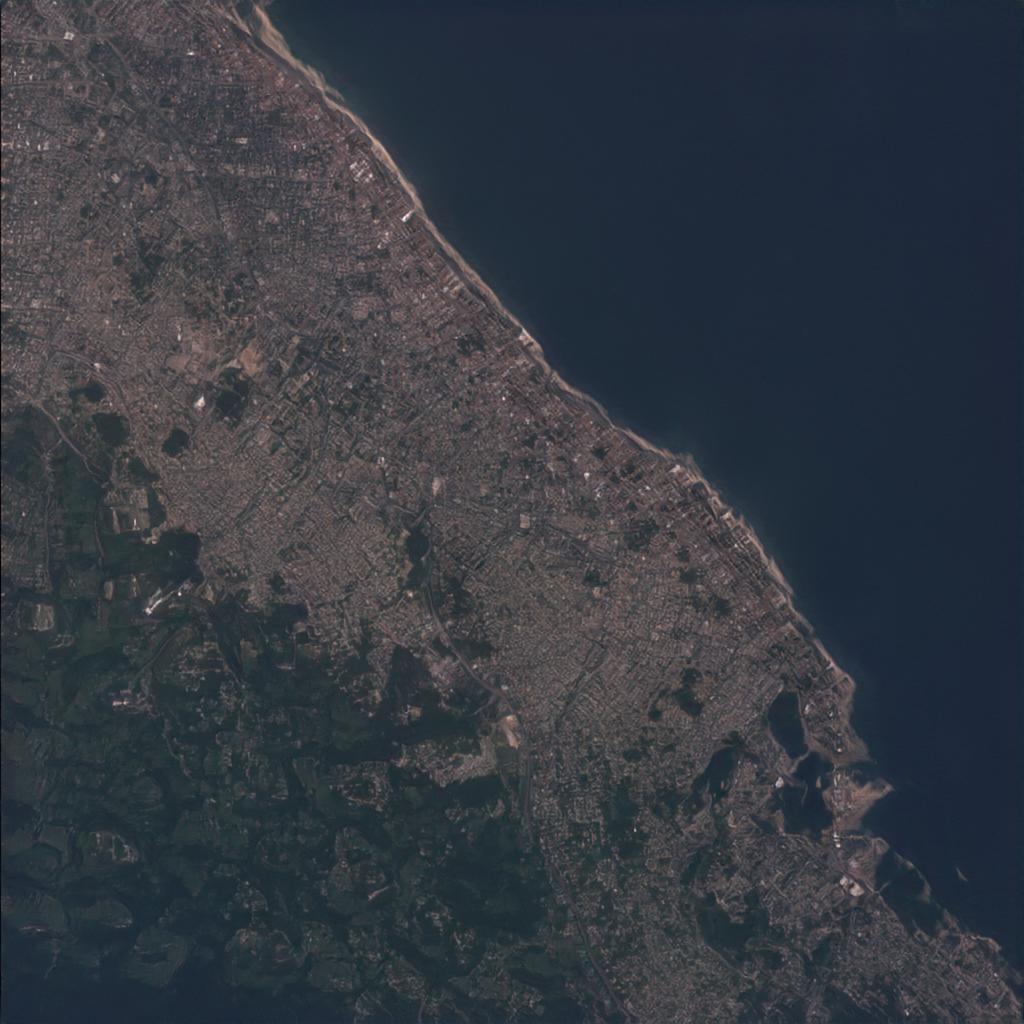}
\includegraphics[width=0.49\columnwidth]{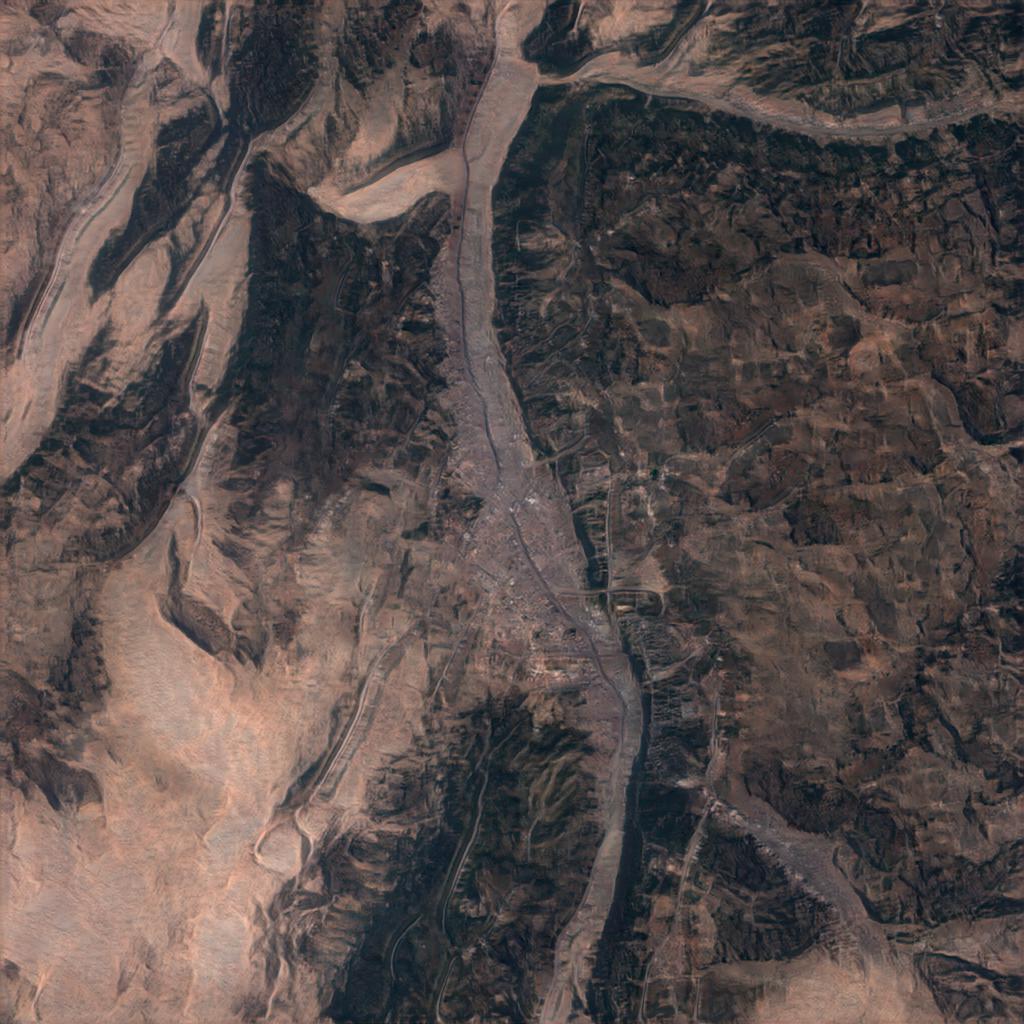}
\caption{This-City-does-not-Exist}
\end{subfigure}
\hfil
\begin{subfigure}{.48\columnwidth}
\label{fig:alps}
\centering
\includegraphics[width=0.49\columnwidth]{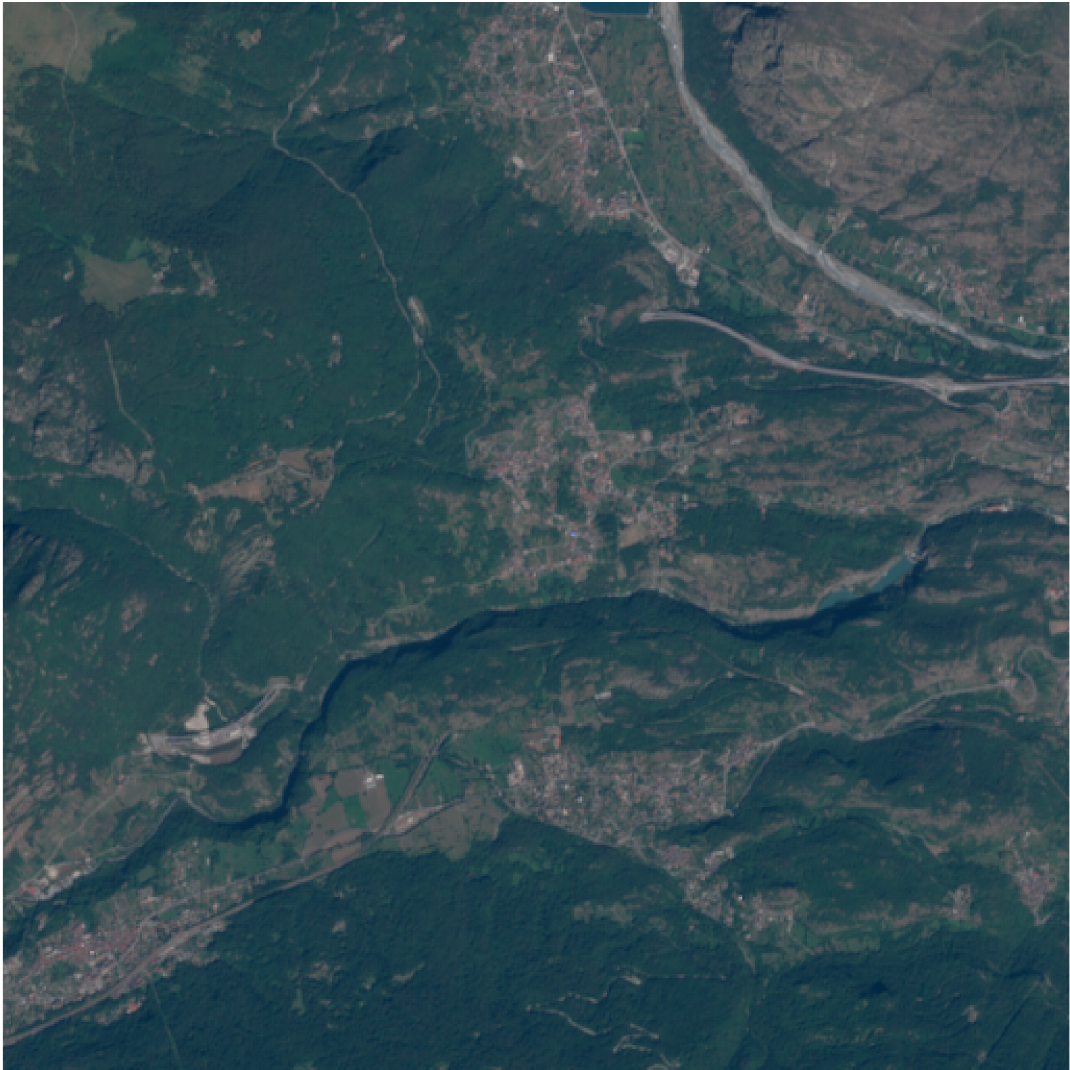}
\includegraphics[width=0.49\columnwidth]{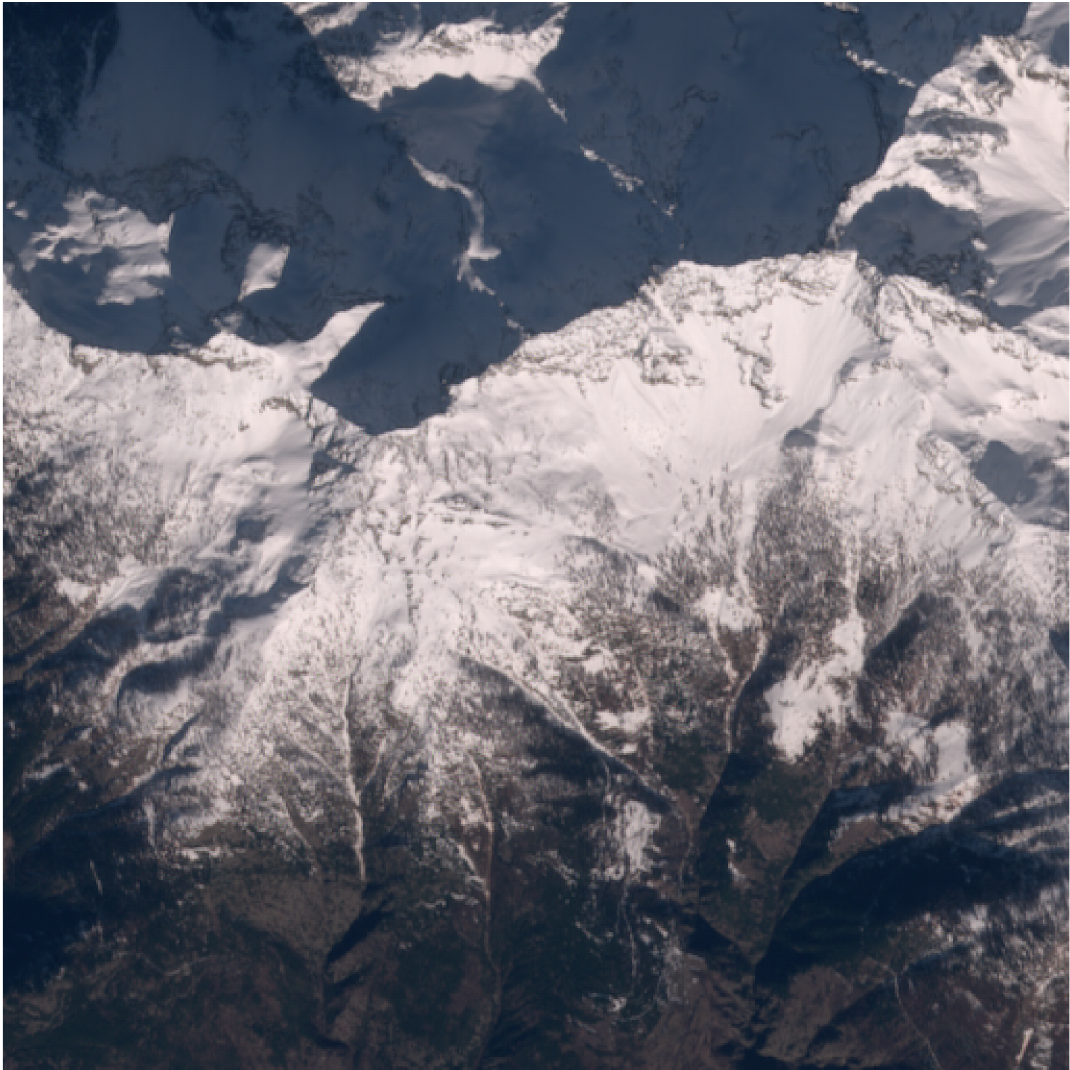}
\caption{Alps dataset.}
\end{subfigure}
\caption{Examples of images taken from this-city-does-not-exit and Alps datasets where for the Alps dataset only RGB bands are shown.}
\label{fig:citydoesntexist_alps}
\end{figure}

This dataset is used to test the ability of the various models to generalize to images generated from unknown architectures and with completely different content. It contains only \gls{gan} images downloaded from \cite{citydoesntexist}. The images have been generated by a styleGAN2 model. We collected 140 images of size $1024\times1024$. The images of this dataset have only 3 bands (RGB). Figure \ref{fig:citydoesntexist_alps}-a, shows two examples of the this-city-does-not-exist dataset.

\section{The VQ VAE 2 one-class classifier}
\label{sec:vqvae2}

In this section, we describe the one-class classifier we have developed to distinguish pristine and \gls{gan} multispectral images. We start with a brief introduction to autoencoders and variational autoencoders, then, we describe the \gls{vqvae2} architecture, which is the one our system relies on.

A neural network $\emph{A}$ trained to reconstruct its input at the output is referred to as an autoencoder \cite{Hinton1993_autoencoders}. The reconstruction is constrained in such a way to prevent learning the identity function. An autoencoder is divided into two main parts:

\begin{itemize}
\item the encoder $\emph{A}_\text{e}$, mapping the input $x$ into a hidden representation $h$ (i.e., $h = \emph{A}_\text{e}(x$)).
\item the decoder $\emph{A}_\text{d}$, reconstructing an approximate version of the input $\tilde{x}$ from the hidden representation (i.e., $\tilde{x} = \emph{A}_\text{d}(h)$).
\end{itemize}
In the case of tensor data, the input can be an image $\X$ and the hidden representation a vector $\mathbf{h}$. The encoder and decoder are trained in tandem to reduce the reconstruction loss, usually a $L_2$ loss term, between the input and output samples.

In \glspl{vae}, the input is encoded into a vectorial representation, and the hidden representation's features are forced to follow a Gaussian distribution, denoted by $\mathcal{N}(f(x); g(x))$, where $f(\cdot)$ denotes the mean and $g(\cdot)$ denotes the variance  of the distribution. A sample of the hidden representation is taken during the decoding stage and utilized as input to the decoder, which produces a reconstructed version of the original input data. When the hidden features are required to follow a Gaussian distribution, the total loss used during training is equal to:
\begin{equation}
\begin{split}
\label{eqn:vaeloss}
\emph{L}(x,\tilde{x}) = {} & \left\lVert x-\tilde{x}\right\rVert^{2}_{2} +
\beta L_{kl}(\mathcal{N}(f(x), g(x)), \mathcal{N}(0, I_d)),
\end{split}
\end{equation}

where the first term is a ``data fidelity term" that measures the difference between the input sample $x$ and the estimated sample $\tilde{x}$, the second term applies a kind of ``regularization" by requiring the network to minimize the Kullback-Leibler divergence $L_{kl}$ between the learned hidden variable distribution and a desired normal distribution $\mathcal{N}(0; I_d)$ (here, $I_d$ is the identity matrix), and $\beta$ is a hyperparameter balancing the two loss terms. The decoder is used to create new images by selecting random samples from the hidden layer after training the \gls{vae}.

\gls{vqvae} \cite{oord2017} is a variant of \gls{vae} \cite{Kingma2019AnIT} that uses vector quantization to learn a discrete latent representation instead of a continuous one. This is done by adding a discrete codebook component to the network that contains the list of vectors with their indices. Then the output of the encoder is compared with all the codebook entries in terms of Euclidean distance, and the code that is closer to the output of the encoder is fed to the decoder. In \gls{vqvae} compared to a \gls{vae}, the priors are learned rather than taken as static input. The combination of a discrete latent representation and an autoregressive prior thus paves the way for the generation of high-quality images, videos, and speech.

\begin{figure}[htbp]
\centering
\includegraphics[clip,width=1\columnwidth]{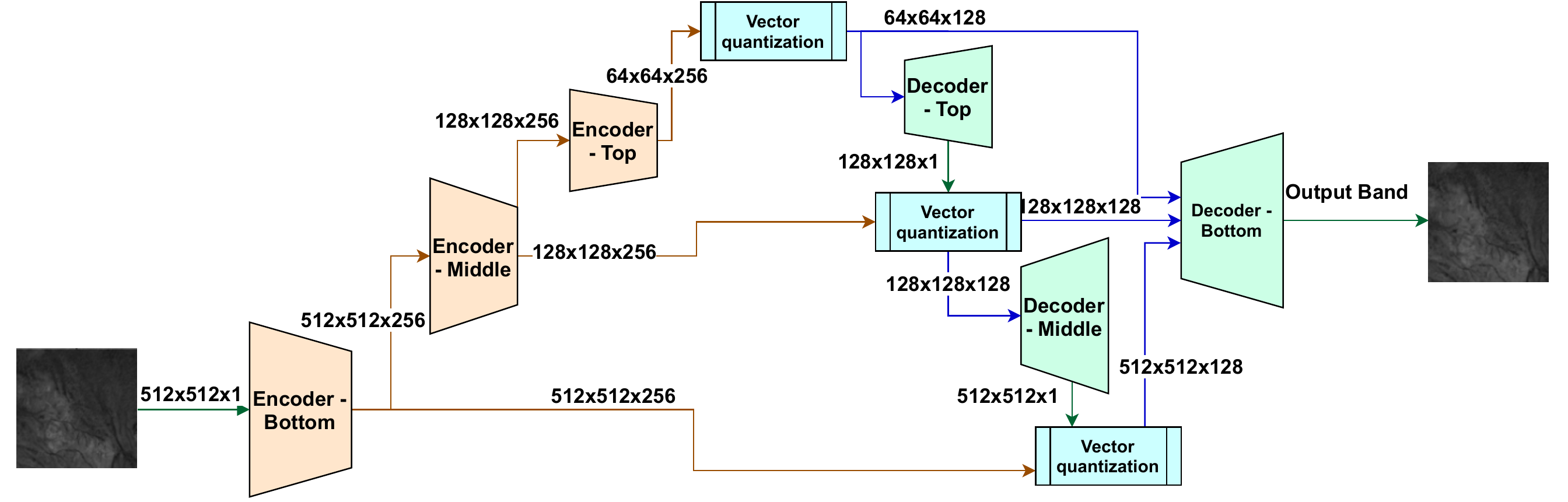}
\caption{VQ-VAE 2 Architecture}
\label{fig:vqvae}
\end{figure}

\gls{vqvae2} \cite{razavi2019} is similar to \gls{vqvae} with the only difference that it uses multi-scale latent maps to increase the resolution of the reconstructed image. In our experiments, we used 3 levels of latent maps and relied only on a static prior instead of an autoregressive prior, since our goal was to detect \gls{gan} images rather than generate them. Figure \ref{fig:vqvae} shows the architecture we used, where we opted for a three-level hierarchy: bottom, middle, and top, with latent space sizes of 512, 128, and 64, respectively.

With regard to the detection of \gls{gan} images, we used the \gls{vqvae2} architecture according to two different modalities. In the first approach, the autoencoder processes all the 13 bands together (the resulting architecture is referred to as \gls{vqvae2}$_{13}$. For the second approach, we trained one model per band (referred to as \gls{vqvae2}$_1$). The reconstruction loss on all bands and the total reconstruction loss calculated on all bands are used as features to be processed by an anomaly detection module, e.g., a one-class \gls{svm}. Based on the experiments we have run (see Section \ref{ssec:vqvae2_eval_detect}),  we decided to use the reconstruction loss directly and detect the \gls{gan} images by applying threshold to the reconstruction error band by band.

\section{Experiments and Results}
\label{sec:eval}

In this section, we evaluate the performance of the proposed \gls{vqvae2} and empirically prove that an off-the-shelf 2-class detector based on EfficientNet-B4 \cite{tan_efficient_2019} lags behind the \gls{vqvae2} in terms of generalization capabilities. Both detectors are trained and tested with the datasets described in Section \ref{sec:datasets}.

\subsection{EfficientNet-B4 Detector}
\label{ssec:eff_eval_detect}

As a baseline 2-class detector to benchmark the performance of our system, we trained a model based on the EfficientNet architecture. The EfficientNet class of networks was proposed as a way to efficiently scale the network depth, width, and resolution based on the input dimensions \cite{tan_efficient_2019}. In our experiments, we used EfficientNet-B4 (eff), with hyper-parameters set as in \cite{tan_efficient_2019}, the only exception being the difference in the input size, which we adapted to match the number of channels our images consist of (13). We built 3 different models by training the networks on the LC dataset (thus fitting to cycleGAN data), the Scandinavian   dataset (adapting the model to distinguish pix2pix images), and a combination of the LC and Scandinavian datasets. The three models were cross-tested on the LC, Scandinavian, and China datasets. We also trained three additional models, this time by removing the down-sampling from the initial layer as suggested by \cite{Gragnaniello_21} to enhance the generalization capabilities of the model. We called these models EfficientNet-B4 with no down (eff\_nodown). Augmentation, including Gaussian blur, random shift, random rotation, and random flip, was applied to train all models.

The results we got on the various datasets are shown in Table \ref{tab:eff_detect_results}, reporting the correct detection probability at a false alarm rate equal to 0.1. For threshold calibration, for each test, we used 100 pristine images from the dataset to be tested. We also obtained other results that we don't report here using different thresholds where we obtained the thresholds from 100 pristine images from the corresponding training dataset. However, the conclusions drawn are the same.

Expectedly, the probability of detection is very good when the datasets used for training and testing are matched, that is when the \gls{gan} images have been generated by the same \gls{gan} model used for training. With regard to generalization, we observe that the eff\_nodown architecture has better generalization capabilities. In any case, the performance drop when the models are tested on images taken from datasets that were not used during training. The best results are obtained by training the detector on images generated by both pix2pix and cycleGAN. Even in this case, though, the performance deteriorates when the detector is tested on the images of the China dataset that was not used during training.


\begin{table}[htb]
\centering
\resizebox{\textwidth}{!}{%
\begin{tabular}{cccccccc}
\hline
\multicolumn{2}{c}{}                                                                                & \multicolumn{6}{c}{\textbf{TEST}}                                                                                                                                                                                                                                                                           \\ \cline{3-8} 
\multicolumn{2}{c}{}                                                                                & \multicolumn{2}{c}{\textbf{LC}}                                                    & \multicolumn{2}{c}{\textbf{Scand}}                                                 & \multicolumn{2}{c}{\textbf{China}}                            \\ \cline{3-8} 
\multicolumn{2}{c}{\multirow{-3}{*}{\textbf{Results}}}                                  & \multicolumn{1}{c}{\textbf{eff\_down}} & \multicolumn{1}{c}{\textbf{eff\_nodown}} & \multicolumn{1}{c}{\textbf{eff\_down}} & \multicolumn{1}{c}{\textbf{eff\_nodown}} & \multicolumn{1}{c}{\textbf{eff\_down}} & \textbf{eff\_nodown} \\ \hline
\multicolumn{1}{c}{{ }}                                 & \textbf{LC}           & {1}               & \multicolumn{1}{c}{\textbf{1}}                & \multicolumn{1}{c}{0.8}              & \multicolumn{1}{c}{1}                   & \multicolumn{1}{c}{0.7}               & 0.82               \\ \cline{2-8} 
\multicolumn{1}{c}{{ }}                                 & \textbf{Scand}        & \multicolumn{1}{c}{0.61}               & \multicolumn{1}{c}{0.73}                   & \multicolumn{1}{c}{\textbf{1}}                & \multicolumn{1}{c}{1}                   & \multicolumn{1}{c}{0.65}              & 0.74                \\ \cline{2-8}
\multicolumn{1}{c}{\multirow{-3}{*}{{ \textbf{TRAIN}}}} & \textbf{LC and Scand} & \multicolumn{1}{c}{1}              & \multicolumn{1}{c}{1}                & \multicolumn{1}{c}{1}              & \multicolumn{1}{c}{1}              & \multicolumn{1}{c}{0.6}              & \textbf{0.86}                \\ \hline
\end{tabular}
}
\caption{EfficientNet-B4 correct detection probability (false alarm rate was set to 0.1).}

\label{tab:eff_detect_results}
\end{table}

\subsection{Vector Quantized Variational Autoencoder 2}
\label{ssec:vqvae2_eval_detect}
To build the \gls{vqvae2} one-class classifiers, we trained a \gls{vqvae2}$_{13}$ model working on all bands together, and 13 models \gls{vqvae2}$_{1}$ models, each working on one different band. The models were trained on 50000 pristine Sentinel 2 level-1C images collected from the Alps, China and land cover datasets. The networks were trained for 100 epochs with early stopping and a batch size equal to 64.

\begin{figure}[htbp]
\begin{subfigure}{.5\columnwidth}
\label{fig:scatterplot_vqvae_13bands}
\centering
\includegraphics[width=0.49\columnwidth]{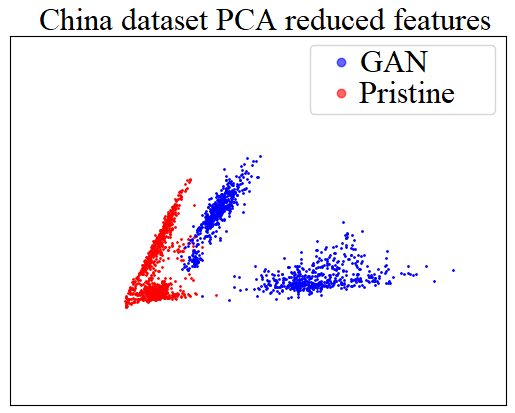}
\includegraphics[width=0.49\columnwidth]{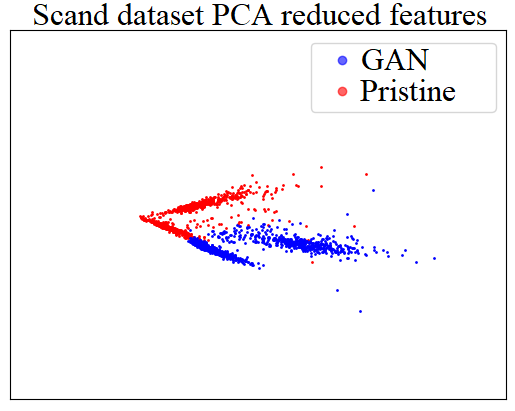}
\caption{VQ-VAE 2 trained on 13 bands}
\end{subfigure}
\hfil
\begin{subfigure}{.5\columnwidth}
\label{fig:scatterplot_vqvae_bandbyband}
\centering
\includegraphics[width=0.49\columnwidth]{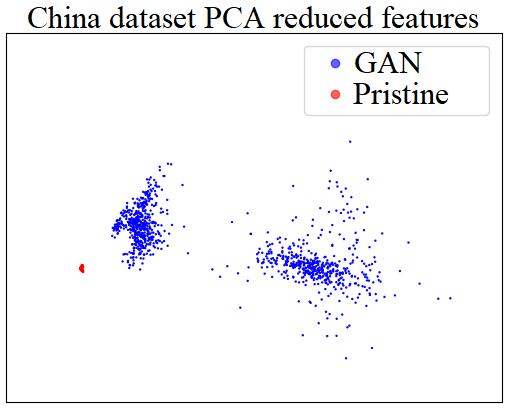}
\includegraphics[width=0.49\columnwidth]{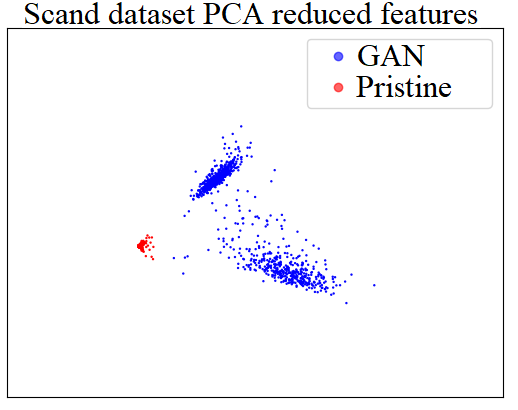}
\caption{VQ-VAE 2 trained band by band}
\end{subfigure}
\caption{Scatter plot after PCA feature reduction for \gls{gan} and pristine images}
\label{fig:scatterplot_vqvae}
\end{figure}

Some initial insights into the discrimination capability of the \gls{vqvae2} models can be obtained by plotting the reconstruction losses obtained by applying the trained autoencoders to the various datasets (see Figure \ref{fig:scatterplot_vqvae}). In order to obtain the scatter plots shown in the Figure, we applied Principle Component Analysis (PCA) to reduce the feature dimensionality to 2. The scatter plots obtained by the models trained on \gls{vqvae2}$_{13}$ and \gls{vqvae2}$_{1}$ are reported for two datasets: the China dataset, whose pristine images were part of the \gls{vqvae2} training dataset, and the Scandinavian dataset, that was never seen by the \gls{vqvae2}. We observe that in both cases, pristine and \gls{gan} images are grouped into well-distinct clusters; however, the features are further apart in the case where the reconstruction losses are taken from the 13 models of \gls{vqvae2}$_{1}$. Hence, the rest of our experiments were carried out by training one model per band.

Given that 13 single-band trained autoencoders, we used the reconstruction error of each autoencoder to detect the \gls{gan} images. To do so, we set the detection threshold by fixing the false alarm rate on 100 pristine images for each testing dataset to 0.1. Similar to the efficientNet-B4 results, we also experimented with obtaining the threshold from the training dataset which in this experiment is obtained from the China and LC datasets. The results, we obtained, are slightly worse for the 10m bands but are comparable for the rest of the bands.


The tests were carried out on the LC, Scand, China and all the datasets mixed together. Table \ref{tab:vqvae_detection_13bands} shows the results we have obtained. The correct detection probability is very good for most of the bands, with some room for improvement on the R (band 4), G (band 3), B (band 2), and NIR (band 8) bands. The results in Table \ref{tab:vqvae_detection_13bands} demonstrate the excellent generalization capability of the \gls{vqvae2}-based detectors, as they are able to detect the \gls{gan} images even in the Scand dataset (whose pristine images were not part of the training dataset) by looking at any of the bands except bands 2 and 3. In particular, the detectors based on bands 7, 8a, 9, 11, and 12 achieve nearly perfect results on all datasets.

\begin{table}[htbp]
\centering
\resizebox{\textwidth}{!}{%
\begin{tabular}{|l|l|l|l|l|l|l|l|l|l|l|l|l|l|}
\hline
\textbf{BANDS}      & \textbf{1} & \textbf{2} & \textbf{3} & \textbf{4} & \textbf{5} & \textbf{6} & \textbf{7} & \textbf{8} & \textbf{8a} & \textbf{9} & \textbf{10} & \textbf{11} & \textbf{12} \\ \hline
\textbf{LC} & 0.94       & 0.88       & 0.9        & 0.55       & 0.83       & 0.92       & 0.99       & 0.7        & 0.99        & 0.98       & 0.99        & 1           & 1           \\ \hline
\textbf{Scand}      & 0.98       & 0.64       & 0.51       & 0.95       & 1          & 0.97       & 0.98       & 0.97       & 0.99        & 1          & 0.99        & 0.97        & 0.99        \\ \hline
\textbf{China}      & 0.99       & 0.98       & 0.99       & 0.98       & 1          & 0.98       & 0.99       & 0.73       & 0.99        & 0.99       & 0.54        & 1           & 1           \\ \hline
\textbf{Mixed data} & 0.85       & 0.86       & 0.6        & 0.81       & 0.92       & 0.95       & 0.98       & 0.7        & 0.99        & 0.98       & 0.85        & 1           & 1           \\ \hline
\end{tabular}}
\caption{Correct detection probability at 0.1 false alarm rates using VQ-VAE 2 for the 13 bands datasets}
\label{tab:vqvae_detection_13bands}
\end{table}

\subsection{Comparison on city-does-not-exist Dataset}
\label{ssec:compare_eval_detect}

To further test the generalization capabilities of the various detectors, we used the "this-city-does-not-exist". The reason for such a choice is twofold. The first reason is that these images were generated by styleGAN 2 which is an architecture that was never used in our experiments. The second reason is that the \gls{vqvae2} autoencoder was trained on the same pristine images of the \gls{gan} architectures, while for the "this-city-does-not-exist" dataset, the \gls{gan} training dataset is unknown to us (and surely different than that we used to train the autoencoder). To carry these experiments, we trained an EfficientNet-B4 no down detector on the LC and Scand datasets, but using only the 3 RGB channels. As for the \gls{vqvae2}, we used the same, single-band, models that had trained before (of course in this case we had to use only the detectors working on the R, G and B bands). The detection thresholds of all detectors were fixed by using a set of pristine Sentinel-2 images, the same 100 we had used before in the case of LC and Scand dataset, this time targeting a False Alarm Rate equal to 0.05. 

The results of these experiments are shown in Table \ref{tab:detection_city}. As we can see, the \gls{vqvae2} detector provides much better results than efficientNet-B4, which is not able to detect properly the images generated from scratch by the styleGAN 2 generator). This is not the case with the \gls{vqvae2} detector that is able to detect the styleGAN 2 images without retraining.

\begin{table}[htbp]
\centering
\resizebox{\textwidth}{!}{%
\begin{tabular}{|c|c|c|c|c|c|}
\hline
\textbf{Metrics}  & \textbf{\begin{tabular}[c]{@{}c@{}}VQ VAE 2\\ (Red)\end{tabular}} & \textbf{\begin{tabular}[c]{@{}c@{}}VQ VAE 2\\ (Blue)\end{tabular}} & \textbf{\begin{tabular}[c]{@{}c@{}}VQ VAE 2\\ ( Green)\end{tabular}} & \textbf{\begin{tabular}[c]{@{}c@{}}EfficientNetB4\\ (trained on 3 bands)\end{tabular}} \\ \hline
\textbf{Accuracy} & 1                                                               & 1                                                                & 0.96                                                                   & 0.72                                                                                                                \\ \hline
\end{tabular}
}
\caption{Correct Detection probability at 0.05 false alarm rate on the "this-city-does-not-exist" dataset.}
\label{tab:detection_city}
\end{table}

\section{Conclusions}
\label{sec:conc}

We have introduced a one-class detector of \gls{gan} multispectral images generated by a variety of DL architectures. The model is based on a \gls{vqvae2} autoencoder and is trained only on pristine images. To the best of our knowledge, this is the first work proposing the use of a one-class classifier to detect 13-band Sentinel-2 level-1C artificially generated images.

We run experiments on images generated by cycleGAN and pix2pix architectures. The results we obtained are particularly promising. In particular, the proposed detector exhibits a superior generalization capability than a baseline 2-class detector based on EfficientNet-B4. To evaluate the generalization capability in extreme conditions, we tested the detector on a small dataset of RGB satellite images generated by styleGAN 2. The proposed detector outperformed the two-class classifier by far. Further work could be done to diversify the generative models that were used to create the satellite images to include additional \gls{gan} and diffusion models.

\backmatter
\bmhead{Acknowledgments}
This material is based on research sponsored by the Defense Advanced Research Projects Agency (DARPA) and the Air Force Research Laboratory (AFRL) under agreement number FA8750-20-2-1004.
The U.S. Government is authorized to reproduce and distribute reprints for Governmental purposes notwithstanding any copyright notation thereon. 
The views and conclusions contained herein are those of the authors and should not be interpreted as necessarily representing the official policies or endorsements, either expressed or implied, of DARPA or AFRL or the U.S. Government.

\section*{Declarations}
\bmhead{Funding}
This material is based on research sponsored by the Defense Advanced Research Projects Agency (DARPA) and the Air Force Research Laboratory (AFRL) under agreement number FA8750-20-2-1004.
\bmhead{Conflict of interest}
The authors declare that they have no conflict of interest.
\bmhead{Ethics approval and Consent to participate}
The authors declare that this research did not require ethical approval or consent to participate since it does not concern human participants or human or animal datasets.
\bmhead{Consent for publication}
The authors of this manuscript consent to its publication. Open Access This article is licensed under a Creative Commons Attribution 4.0 International License, which permits use, sharing, adaptation, distribution and reproduction in any medium or format, as long as you give appropriate credit to the original author(s) and the source, provide a link to the Creative Commons licence, and indicate if changes were made. The images or other third party material in this article are included in the article’s Creative Commons licence, unless indicated otherwise in a credit line to the material. If material is not included in the article’s Creative Commons licence and your intended use is not permitted by statutory regulation or exceeds the permitted use, you will need to obtain permission directly from the copyright holder. To view a copy of this licence, visit http://creativecommons.org/licenses/by/4.0/.
\bmhead{Authors' contributions} Lydia Abady wrote the first draft of this manuscript and ran the experiments. Prof. Mauro Barni provided feedback and guidance in all the steps of the this research and was involved in reviewing and improving the manuscript. Giovanna Maria Dimitri provided feedback for research and provided reviews for the manuscript. The research idea developed was an equal contribution between Lydia Abady and Prof. Barni.
\bmhead{Code availability} NA
\bmhead{Availability of data and material} NA

\bibliography{refs.bib}
\end{document}

%% file: notation_helper.tex
\newacronym{sota}{SOTA}{State of the Art}
\newacronym{nn}{NN}{Neural Network}
\newacronym{dnn}{DNN}{Deep Neural Network}
\newacronym{ssl}{SSL}{Semi-Supervised Learning}
\newacronym{aviris}{AVIRIS}{Airborne Visible/Infrared Imaging Spectrometer}
\newacronym{hsi}{HSI}{Hyper-Spectral Imagery}
\newacronym{cnn}{CNN}{Convolutional Neural Network}
\newacronym{gan}{GAN}{Generative Adversarial Network}
\newacronym{cgan}{cGAN}{Conditional Generative Adversarial Network}
\newacronym{nir}{NIR}{Near Infrared}
\newacronym{sgd}{SGD}{Stochastic Gradient Descent}
\newacronym{lr}{lr}{Learning Rate}
\newacronym{oa}{OA}{Overall Accuracy}
\newacronym{msi}{MSI}{Multi-Spectral Imagery}
\newacronym{gsd}{GSD}{Ground Sampling Distance}
\newacronym{toa}{TOA}{Top-of-Atmosphere}
\newacronym{boa}{BOA}{Bottom-of-Atmosphere}
\newacronym{wgan-gp}{WGAN-GP}{Wasserstein GAN  with Gradient Penalty loss}
\newacronym{msg}{MSG}{Multi-Scale Gradients}
\newacronym{rmsprop}{RMSProp}{Root Mean Squared Propagation}
\newacronym{nicegan}{NICE-GAN}{No Independent Component for Encoding GAN}
\newacronym{snap}{SNAP}{Sentinel Application Platform}
\newacronym{vv}{VV}{vertical transmit and vertical receive}
\newacronym{hh}{HH}{horizontal transmit and horizontal receive}
\newacronym{slc}{SLC}{Simple Look Complex}
\newacronym{grd}{GRD}{Ground Range Detected}
\newacronym{mstar}{MSTAR}{Moving and Stationary Target Acquisition and Recognition}
\newacronym{ocn}{OCN}{Ocean}
\newacronym{vh}{VH}{vertical transmit and horizontal receive}
\newacronym{hw}{HV}{horizontal transmit and vertical receive}
\newacronym{psnr}{PSNR}{peak signal-to-noise ratio}
\newacronym{ssim}{SSIM}{structural similarity}
\newacronym{mae}{MAE}{mean absolute error}
\newacronym{mape}{MAPE}{mean absolute percentage error}
\newacronym{msa}{MSA}{mean spectral angle}
\newacronym{svm}{SVM}{Support Vector Machine}
\newacronym{sar}{SAR}{Synthetic Aperture Radar}
\newacronym{wv}{WV}{Wave}
\newacronym{iw}{IW}{Interferometric Wide swath}
\newacronym{ew}{EW}{Extra Wide swath}
\newacronym{sm}{SM}{Stripmap}
\newacronym{knn}{KNN}{K-Nearest Neighbors}
\newacronym{km}{KM}{K-Means}
\newacronym{gmm}{GMM}{Gaussian Mixture Models}
\newacronym{dbn}{DBN}{Deep Belief Network}
\newacronym{rf}{RF}{Random Forest}
\newacronym{bce}{BCE}{Binary Cross Entropy}
\newacronym{prnu}{PRNU}{Photo Response Non-Uniformity}
\newacronym{ccd}{CCD}{Charge-Coupled Devices}
\newacronym{cmos}{CMOS}{Complementary Metal-Oxide Semiconductor}
\newacronym{svdd}{SVDD}{Support Vector Data Descriptor}
\newacronym{rbm}{RBM}{Restricted Boltzmann Machine}
\newacronym{naun}{NAU-N}{Nested Attention U-Net}
\newacronym{eo}{EO}{Electro-Optical}
\newacronym{lwir}{LWIR}{Long-Wave Infra-Red}
\newacronym{esa}{ESA}{European Space Agency}
\newacronym{usgs}{USGS}{United States Geological Survey}
\newacronym{unoosa}{UNOOSA}{United Nations - Office for Outer Space Affairs}
\newacronym{gcp}{GCP}{Ground Control Points}
\newacronym{ncc}{NCC}{Normalized cross-correlation}
\newacronym{cc}{CC}{correlation coefficient}
\newacronym{sam}{SAM}{spectral angle mapping}
\newacronym{rmse}{RMSE}{root mean squared error}
\newacronym{sift}{SIFT}{Scale Invariant Feature Transform}
\newacronym{brisk}{BRISK}{Binary Robust Invariant Scalable Key}
\newacronym{progan}{ProGAN}{Progressive Generative Adversarial Network}
\newacronym{vae}{VAE}{Variational Autoencoder}
\newacronym{dem}{DEM}{Digital Elevation Models}
\newacronym{spade}{SPADE}{spatially adaptive normalization}
\newacronym{iou}{IoU}{intersection over union}
\newacronym{nwp}{NWP}{numerical weather prediction}
\newacronym{mdn}{MDN}{mixture density network}
\newacronym{tsne}{t-SNE}{t-distributed stochastic neighbor embedding}
\newacronym{cyclegan}{CycleGAN}{Cycle Generative Adversarial Network}
\newacronym{stgan}{STGAN}{Spatio-Temporal Generative Adversarial Network}
\newacronym{swir}{SWIR}{Short-Wave InfraRed}
\newacronym{saab}{SAAB}{Subspace Approximation with Adjusted Bias}
\newacronym{vqvae2}{VQ-VAE 2}{Vector Quantized Variational Autoencoder 2}
\newacronym{vqvae}{VQ-VAE}{Vector Quantized Variational Autoencoder}

\newacronym{dl}{DL}{Deep Learning}

\def\X{\mathbf{X}}

\definecolor{purple}{RGB}{165,0,255}